\documentclass[number,preprint,3p]{elsarticle}

\usepackage{nicefrac}
\usepackage{color}
\usepackage{graphicx}
\usepackage{subcaption}
\usepackage{algorithm}
\usepackage{bm}
\usepackage[colorlinks]{hyperref}
\usepackage{amssymb}
\usepackage{amsthm}
\usepackage{amsmath}
\usepackage{amssymb}
\usepackage{mathtools}
\usepackage{dsfont}
\usepackage{booktabs}
\usepackage{url}
\usepackage{scrextend}
\usepackage{epstopdf}
\usepackage{float}
\usepackage{tcolorbox}
\usepackage{tabularx}
\usepackage{multirow}
\usepackage{tablefootnote}
\usepackage[perpage]{footmisc}
\usepackage{bbm}
\usepackage{lineno}
\usepackage{soul}

\DeclareUnicodeCharacter{0301}{\'{e}}


\makeatletter
\gdef\urlauthor#1#2{\g@addto@macro\@elsuads{\let\corref\@gobble%
     \def\@@tmp{#1}\raggedright\eadsep
     {\ttfamily\url{\expandafter\strip@prefix\meaning\@@tmp}}\space(#2)%
     \def\eadsep{\unskip,\space}}%
}
\gdef\emailauthor#1#2{\stepcounter{ead}%
     \g@addto@macro\@elseads{\raggedright%
      \let\corref\@gobble\def\@@tmp{#1}%
      \eadsep{\ttfamily\href{mailto:\expandafter\strip@prefix\meaning\@@tmp}{\expandafter\strip@prefix\meaning\@@tmp}}
      (#2)\def\eadsep{\unskip,\space}}%
}
\makeatother
\biboptions{numbers,sort&compress,square}
\journal{arXiv}

\begin{document}
    \begin{frontmatter}
    \renewcommand{\thefootnote}{\fnsymbol{footnote}}
    \title{Deep learning-based modularized loading protocol for parameter estimation of Bouc-Wen class models}
    \author[1]{Sebin Oh}
    \author[2]{Junho Song\corref{cor1}}
    \ead{junhosong@snu.ac.kr}  
    \author[3]{Taeyong Kim\corref{cor1}}
    \ead{taeyongkim@ajou.ac.kr}  
    \cortext[cor1]{Corresponding author}
    \address[1]{Department of Civil and Environmental Engineering, University of California, Berkeley, CA, United States}
    \address[2]{Department of Civil and Environmental Engineering, Seoul National University, Seoul, Republic of Korea}
    \address[3]{Department of Civil Systems Engineering, Ajou University, Suwon, Republic of Korea}
    \begin{abstract}
            \noindent
            This study proposes a modularized deep learning-based loading protocol for optimal parameter estimation of Bouc-Wen (BW) class models. The protocol consists of two key components: optimal loading history construction and CNN-based rapid parameter estimation. Each component is decomposed into independent sub-modules tailored to distinct hysteretic behaviors-basic hysteresis, structural degradation, and pinching effect-making the protocol adaptable to diverse hysteresis models. Three independent CNN architectures are developed to capture the path-dependent nature of these hysteretic behaviors. By training these CNN architectures on diverse loading histories, minimal loading sequences, termed \textit{loading history modules}, are identified and then combined to construct an optimal loading history. The three CNN models, trained on the respective loading history modules, serve as rapid parameter estimators. Numerical evaluation of the protocol, including nonlinear time history analysis of a 3-story steel moment frame and fragility curve construction for a 3-story reinforced concrete frame, demonstrates that the proposed protocol significantly reduces total analysis time while maintaining or improving estimation accuracy. The proposed protocol can be extended to other hysteresis models, suggesting a systematic approach for identifying general hysteresis models. The source codes, data, and trained CNN models are available for download at [URL that will be available once the paper is accepted].
    \end{abstract}
    
        
    \end{frontmatter}
    
    \renewcommand{\thefootnote}{\fnsymbol{footnote}}

    \section{Introduction} \label{section:Intro}
    \noindent 
    Seismic design and assessment of structural systems typically begin with the estimation of responses under a set of ground motions. When estimating structural response, it is necessary to numerically model the hysteretic behavior based on the understanding of its characteristics. The hysteretic behavior represents the path-dependent nonlinear relationship between a structure's deformation and the force it resists \cite{Ismail2009TheSurvey}. Among various hysteresis models, Bouc-Wen (BW) class models are widely adopted owing to their ability to capture complex hysteretic phenomena, including structural degradation and the pinching effect. These models provide a smooth curve of the force-deformation relationship defined by nonlinear differential equations \cite{Song2006GeneralizedHysteresis, Kim2021NondimensionalizedMonitoring,Bouc1967ForcedHysteresis,Ismail2009TheSurvey, Ma2004ParameterHysteresis,Kim2019ResponseNetworks}. Numerous methods have been developed to estimate the parameters of the BW class models for characterizing the dynamic properties of structural systems \cite{Niola2019NonlinearIsolators,Charalampakis2008IdentificationAlgorithm,Xie2020SeismicModel,Wei2022HystereticModel}.

    To estimate the parameters of BW class model, a predefined loading history, i.e., displacement step, is required. In accordance with design codes and guidelines, the loading history is typically characterized by gradually increasing amplitudes history over successive cycles \cite{ACICommittee3742013GuideLoads,ATC-241992GuidelinesStructures,FEMA2007InterimComponents, Krawinkler2001DevelopmentStructures}. Once the hysteresis curve of a target structural system is obtained from this loading history, optimization techniques are employed to identify a parameter set that minimizes the difference between the hysteretic curve of the structural system and the one reproduced using the parameter set \cite{Ismail2009TheSurvey}. However, two main factors, in general, hamper the reliable parameter estimation: (1) the limited investigation of the loading histories widely used for fitting purpose, and (2) the functional redundancy of the BW class models.
    
    The aforementioned loading histories, designed to simulate realistic seismic damage, are effective for parameter estimation in certain aspects, e.g., elastic range or hardening behavior. However, their overall effectiveness has not been thoroughly examined, leaving uncertainty about whether they are sufficient or redundant for accurate parameter identification. \citet{Oh2023BoucWenAnalysis} demonstrated that conventional loading histories are insufficient for identifying parameters associated with more complex hysteretic behavior, such as acute deterioration and the pinching effect, calling for the need to investigate a wide variety of loading histories. Yet, examining multiple loading histories demands significant resources, both in experimental and computational aspects \cite{Zhou2016ExperimentalLoading, Ozdemir2018VariationsLoading, Gatto2003EffectsShearwalls}. Several research efforts have aimed at developing numerical hysteresis models that are independent of input loading history to circumvent this issue \cite{Lee2021AnProtocols, Li2012ModelingAlgorithm}. Nevertheless, these models have not yet become the standard for capturing such complex relationships due to limited validation by dynamic analyses and restricted applicability to general civil structures.
    
    The functional redundancy of BW class models further complicates unique identification of model parameters. This redundancy refers to the existence of different parameter sets producing equivalent hysteresis curves. There are two main factors that contribute to this redundancy: (1) some parameters in the BW model are functionally interrelated and can compensate for each other, and (2) nonlinearity of differential equations of the BW model. To address this issue, some researchers have attempted to use different objective functions for parameter identification, depending on the focus of their study \cite{Zsarnoczay2020UsingMethods, Ortiz2013IdentificationAlgorithms, Ismail2009TheSurvey}. However, this approach has not achieved sufficient universality in applications to the seismic analysis of general structural systems or specimens. 

    To address this research need, this study proposes a deep learning-based loading protocol for efficient and reliable parameter estimation of BW class models. A loading protocol refers to the series of processes for parameter estimation, which encompasses more than just the loading history. Among the wide spectrum of deep learning methods, convolutional neural networks (CNNs) are adopted owing to their ability to capture spatial correlations between data points \cite{Geron2017Hands-onSystems, Khan2018AVision, Alzubaidi2021ReviewDirections, He2016DeepRecognition, Deng2009Imagenet:Database} because the path-dependent nature of the hysteretic behavior can be effectively interpreted as spatial correlations between adjacent points in the data \cite{Kim2019ResponseNetworks}. The proposed loading protocol focuses on three key hysteresis categories commonly observed in civil structures: (1) basic hysteresis, (2) structural degradation, and (3) pinching effect. By utilizing CNN models as surrogates, a comprehensive examination of the impact of loading histories on parameter estimation in each category can be conducted. The introduction of deep learning models also avoids the need to determine a specific objective function by directly considering the difference between true and predicted parameter values as the objective function.
    
\begin{figure}[H]
        \centering
        \includegraphics[width=1.0\linewidth]{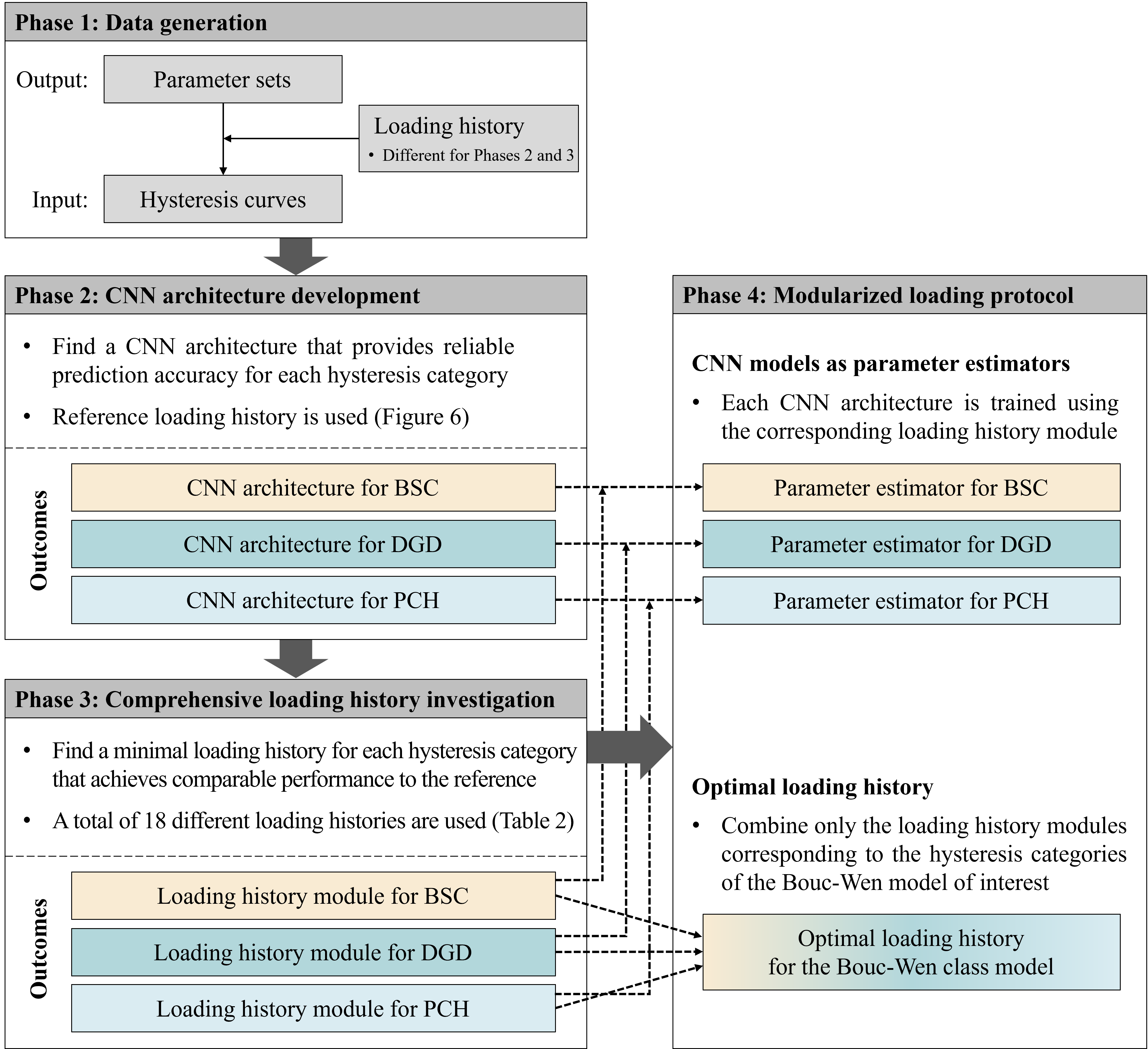}
        \caption{\textbf{Overall framework to develop a modularized loading protocol. BSC, DGD, and PCH denote three hysteresis categories: basic hysteresis, structural degradation, and the pinching effect.}}
        \label{fig:overall_framework}
    \end{figure}

    The overall procedure for developing the loading protocol using CNN models is illustrated in Figure \ref{fig:overall_framework}, where BSC, DGD, and PCH denote three hysteresis categories: basic hysteresis, structural degradation, and the pinching effect, respectively. First, we develop a CNN architecture for each hysteresis category that effectively captures the behavior. The conventional loading history suggested in the guidelines is used as the reference loading history to develop the CNN architecture. The CNN architectures are then trained on a comprehensive set of loading histories to identify loading histories that provide parameter estimation performance comparable to, or better than, the reference loading history. Among the identified histories, the one with the smallest size is selected and termed a \textit{loading history module}. The loading history module is obtained for each hysteresis category. Moreover, the CNN architectures trained on the identified loading history modules serve as rapid parameter estimators. The loading history module and the parameter estimator, developed independently for each hysteresis category, make the overall procedure modular. Finally, the optimal loading history is constructed based on the loading history modules, tailored to the hysteretic characteristics of the BW class model under consideration.

    The remainder of the paper is organized as follows: Section \ref{section:mBWBN} details the m-BWBN model \cite{Kim2023DeepPinching}, which is selected as a representative BW class model to illustrate the protocol development procedure. Section \ref{section:CNNmodels} outlines the CNN architecture developed to capture the hysteretic behavior. Section \ref{section:loading_history} explores the identification of loading history modules, along with the proposed optimal loading histories for four different BW class model. Section \ref{section:numerical_investigations} demonstrates the proposed loading protocol and its applications to two different structural systems: a 3-story steel moment frame and a 3-story reinforced concrete frame. Finally, Section \ref{section:conclusions} concludes with a discussions on limitations and future research directions to clarify and facilitate the applicability of the proposed loading protocol.
    
    \section{Review: modified Bouc-Wen-Baber-Noori model}\label{section:mBWBN}
    \noindent
    \citet{Kim2023DeepPinching} proposed the modified Bouc-Wen-Baber-Noori (m-BWBN) model to enhance the practicality of the Bouc-Wen-Baber-Noori (BWBN) model \cite{Baber1985RANDOMSYSTEMS, baber1986modeling}, by rearranging the mathematical formulation to explicitly control the yield strength parameter. The following section provides a high-level overview  of the m-BWBN model along with a brief examination of each parameter.

    \subsection{Mathematical formulation} \label{section:mBWBN_formulation}
    \noindent
    The BW class models, in general, define the resisting force $f_s$ in terms of  linear and nonlinear components as follows:
    \begin{equation}\label{eq:resisting_force_original}
        f_s(u,z)=\alpha k_0 u + (1-\alpha)k_0z
    \end{equation}
    where $u$ represents the displacement; $z$ denotes the hysteretic displacement with dimensions of length; $\alpha$ is the post-to-pre-yield stiffness ratio; and $k_0$ is the initial stiffness. To explicitly incorporate the yield strength $F_y$, the m-BWBN model uses a modified formulation
    \begin{equation}\label{eq:resisting_force_mBWBN}
        f_s(u,z)=\alpha k_0 u + (1-\alpha)F_yz,
    \end{equation}
    where, $z$ becomes a \textit{dimensionless} variable whose time derivative is defined by the following differential equation:
    \begin{equation}\label{eq:z_ode}
        \dot{z}=\frac{h(z,\varepsilon_n)}{\eta(\varepsilon_n)}\dot{u}\left(1-\lvert z \rvert^n(\beta\text{sgn}(\dot{u}z)+\gamma)\nu(\varepsilon_n)\right)\frac{1}{u_y}
    \end{equation}
    where $\dot{u}$ is the time derivatives of  $u$; $\varepsilon_n$ is the normalized cumulative hysteretic energy; $h(\cdot)$ is the pinching function; $\eta(\cdot)$ is the stiffness degradation function; $\nu(\cdot)$ is the strength degradation function; $n$ is a parameter controlling the sharpness of the yielding transition; $\beta$ and $\gamma$ are the basic shape parameters; sgn$(\cdot)$ is the signum function; and $u_y=\frac{F_y}{k_0}$ is the yield displacement. The normalized cumulative hysteretic energy $\varepsilon_n$ is calculated from the following differential equation:
    \begin{equation}\label{eq:epsilon_n}
        \dot{\varepsilon}_n=(1-\alpha)F_yz\dot{u}.
    \end{equation}
    
    Moreover, the functions describing the hysteretic behaviors in Eq.\eqref{eq:z_ode}, $\eta(\varepsilon_n)$, $\nu(\varepsilon_n)$, and $h(z,\varepsilon_n)$, are defined as
    \begin{align}
        \eta(\varepsilon_n)&=1+\delta_\eta\varepsilon_n \label{eq:eta} \\ 
        \nu(\varepsilon_n) &=1+\delta_\nu\varepsilon_n \label{eq:nu} \\ 
        h(z,\varepsilon_n) &=1-\zeta_1(\varepsilon_n)\exp\left(-\frac{(z\cdot\text{sgn}(\dot{u})-qz_u)^2}{\zeta_2^2(\varepsilon_n)}\right) \label{eq:pinching}
    \end{align}
    in which $\delta_\eta$ and $\delta_\nu$ determine the rates of degradation in stiffness and strength, respectively; $q$ controls the initiation of pinching; $z_u=\left(\frac{1}{\nu(\varepsilon_n)(\beta+\gamma)}\right)^\frac{1}{n}$ is the ultimate value for $z$; and $\zeta_1(\varepsilon_n)$ and $\zeta_2(\varepsilon_n)$ control the progress of pinching, given by
    \begin{align}
        \zeta_1(\varepsilon_n) &= \zeta_0 \left(1-\exp\left(-p\varepsilon_n \right) \right) \label{eq:zeta1} \\
        \zeta_2(\varepsilon_n) &= \left( \psi+\delta_\psi \varepsilon_n \right) \left( \lambda+\zeta_1(\varepsilon_n) \right) \label{eq:zeta2}
    \end{align}
    where $\zeta_0$ indicates the total slip; $p$ contributes to the initial drop rate of the stiffness due to the pinching; $\psi$ and $\lambda$ are the parameters controlling the evolution of $\zeta_2$; and $\delta_\psi$ affects the rate of pinching.

    In summary, the m-BWBN model comprises a total of 14 parameters. However, by adopting the constraint $\beta+\gamma=1$ as suggested by \citet{Kim2023DeepPinching}, the number of parameters reduces to 13. Among these, five parameters ($\alpha$, $k_0$, $F_y$, $\beta$ or $\gamma$, and $n$), referred to as basic parameters (BSC), form the overall shape of the hysteresis curve. The initial stiffness and the yield strength are often normalized by mass, making Eq.\eqref{eq:resisting_force_mBWBN} in units of acceleration. Accordingly, the initial stiffness $k_0$ is commonly replaced with the natural period $T$ using the relationship $k_0=\left(\frac{2\pi}{T}\right)^2$, where $g$ denotes gravitational acceleration. Adopting this adjustment, this study defines the BSC parameter sets as $\alpha$, $T$, $F_y$, $\beta$, and $n$. Two parameters ($\delta_\eta$ and $\delta_\nu$), classified as degradation parameters (DGD), determine the structural degradation, while the remaining six ($\zeta_0$, $p$, $q$, $\psi$, $\delta_\psi$, $\lambda$), categorized as pinching parameters (PCH), control the degree of pinching observed in the hysteresis curve. This study introduces a loading history module tailored to these three hysteresis categories.     
    
    \subsection{Parameter analysis}\label{section:parameter_analysis}
    \noindent    
    Since the BW class models are governed by nonlinear differential equations, there is a possibility of generating unrealistic and non-physical parameter hysteresis curve \cite{Oh2023BoucWenAnalysis, Kim2023DeepPinching, Zsarnoczay2020UsingMethods}. Moreover, since different parameter sets can produce similar hysteresis curves, it is  important to define the parameter ranges that accurately represent real-world civil structures. To address this, this study adopts the parameter bounds of the m-BWBN model provided by \citet{Kim2023DeepPinching}. The bounds were derived using a genetic algorithm based on 416 sets of experimental data from various types of reinforced concrete columns. A summary of such bounds, along with a brief description of their role, is presented in Table \ref{table:params_mBWBN}. Here, $g$ denotes gravitational acceleration, the yield strength $F_y$ is normalized by mass, and the natural period $T$ is adopted instead of the initial stiffness $k_0$.

    \begin{table}[H]
    \caption{\textbf{Roles, bounds, and assumed distributions for the parameters of the m-BWBN model, grouped by three hysteresis categories, where $g$ denotes gravitational acceleration.}}
    \label{table:params_mBWBN}
    \vspace{-0.2cm}
    \centering
    \begin{tabularx}{\linewidth}{@{}l>{\hsize=0.8\hsize}X>{\hsize=2.0\hsize}l>{\hsize=1.2\hsize}X>{\hsize=1.0\hsize}l@{}}
        \toprule[1.0pt]
        \textbf{Category}       & \textbf{Parameter}    & \textbf{Role}                     & \textbf{Bounds}                   & \textbf{Distribution}                     \\ \midrule[1.0pt]
        \multirow{5}{*}{BSC}    & $T$                   & Natural period                 & $0.05$ s $\leq T \leq 5$ s        & Uniform                                   \\ 
                                & $F_y$                 & Normalized yield strength         & $0.05g \leq F_y \leq 1.5g$        & Uniform                                   \\ 
                                & $\alpha$              & Post-to-pre-yield stiffness ratio & $0 \leq \alpha \leq 0.5$          & Truncated normal                          \\ 
                                & $\beta$ (or $\gamma$) & Basic shape control               & $0.1 \leq \beta \leq 0.9$         & Uniform                                   \\ 
                                & $n$                   & Sharpness of yield                & $1 \leq n \leq 5$                 & Uniform                                   \\ \midrule
        \multirow{2}{*}{DGD}    & $\delta_{\nu}$        & Strength degradation rate         & $0 \leq \delta_{\nu} \leq 0.36$   & Truncated normal                          \\ 
                                & $\delta_{\eta}$       & Stiffness degradation rate        & $0 \leq \delta_{\eta} \leq 0.39$  & Truncated normal                          \\ \midrule
        \multirow{6}{*}{PCH}    & $\zeta_0$             & Measure of total slip             & $0 \leq \zeta_0 \leq 1$           & \multirow{6}{*}{Truncated joint normal}   \\ 
                                & $p$                   & Pinching slope                    & $0 \leq p \leq 1.38$              &                                           \\ 
                                & $q$                   & Pinching initiation               & $0.01 \leq q \leq 0.43$           &                                           \\ 
                                & $\psi$                & Pinching magnitude                & $0.1 \leq \psi \leq 0.85$         &                                           \\ 
                                & $\delta_{\psi}$       & Pinching rate                     & $0 \leq \delta_{\psi} \leq 0.09$  &                                           \\ 
                                & $\lambda$             & Pinching severity                 & $0.01 \leq \lambda \leq 0.8$      &                                           \\ 
        \bottomrule[1.0pt]
    \end{tabularx}
    \end{table}
    
\begin{figure}[H]
        \centering
        \includegraphics[height=7cm]{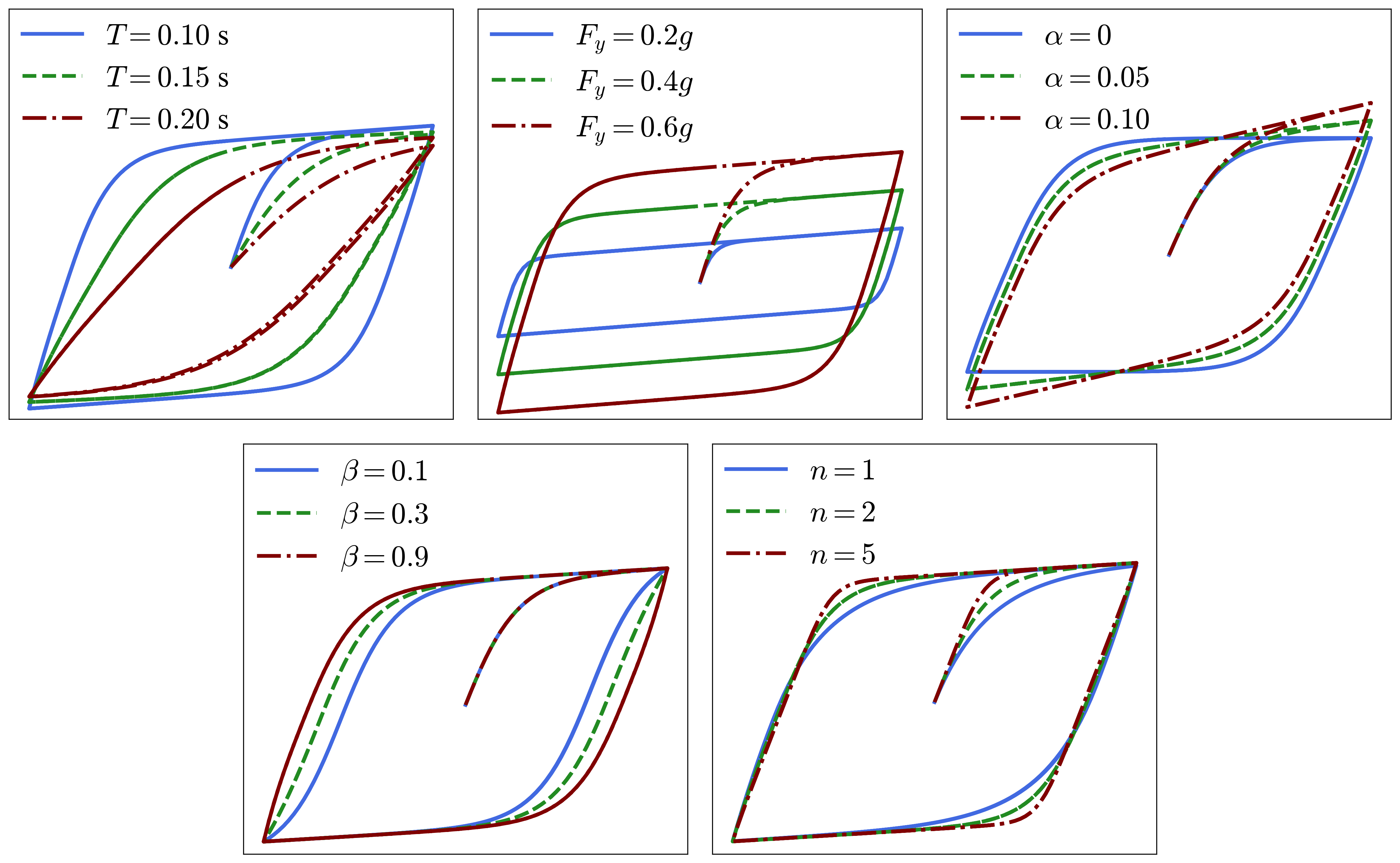}
        \caption{\textbf{Variations of the hysteresis curves for three different values of the BSC parameters.}}
        \label{fig:param_analy_bsc}
    \end{figure}

\begin{figure}[H]
        \centering
        \includegraphics[height=3.5cm]{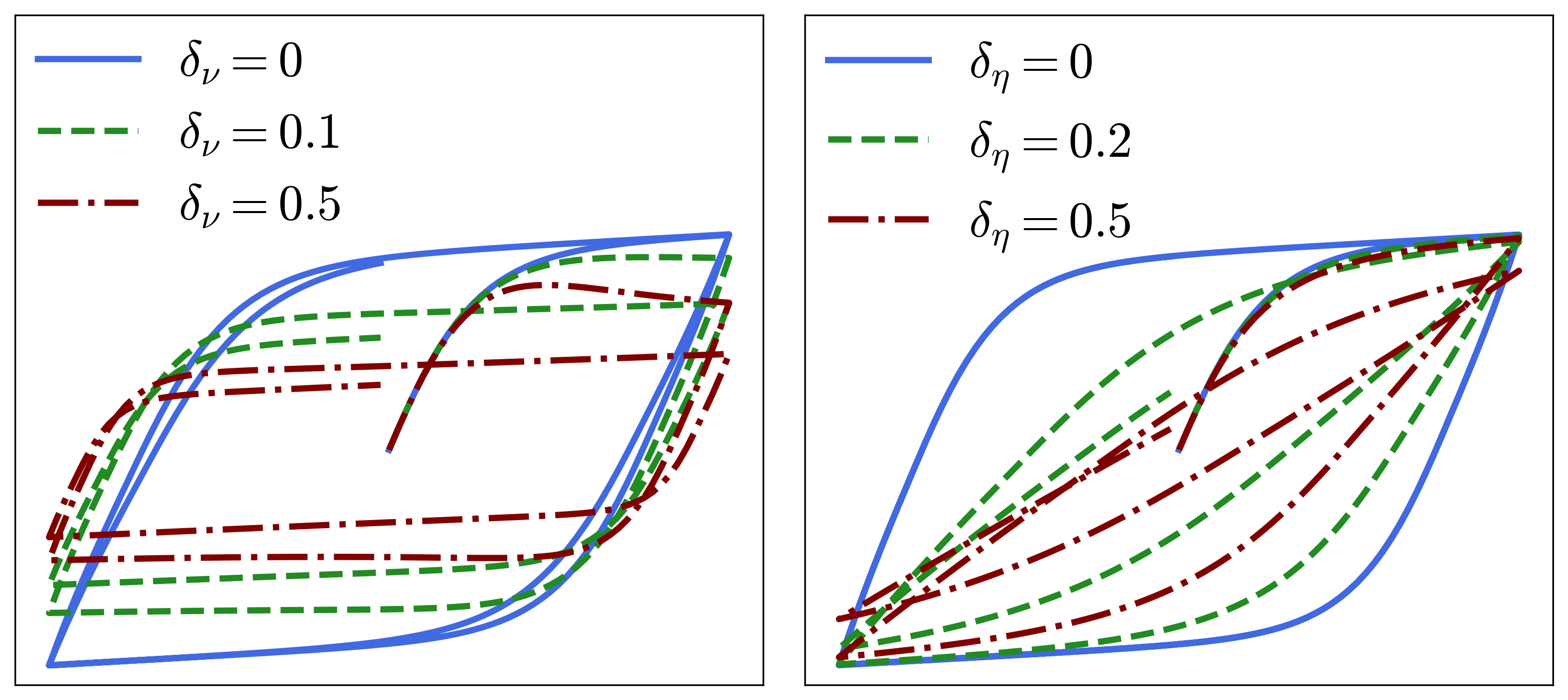}
        \caption{\textbf{Variations of the hysteresis curves for three different values of the DGD parameters.}}
        \label{fig:param_analy_dgd}
    \end{figure}

\begin{figure}[H]
        \centering
        \includegraphics[height=7cm]{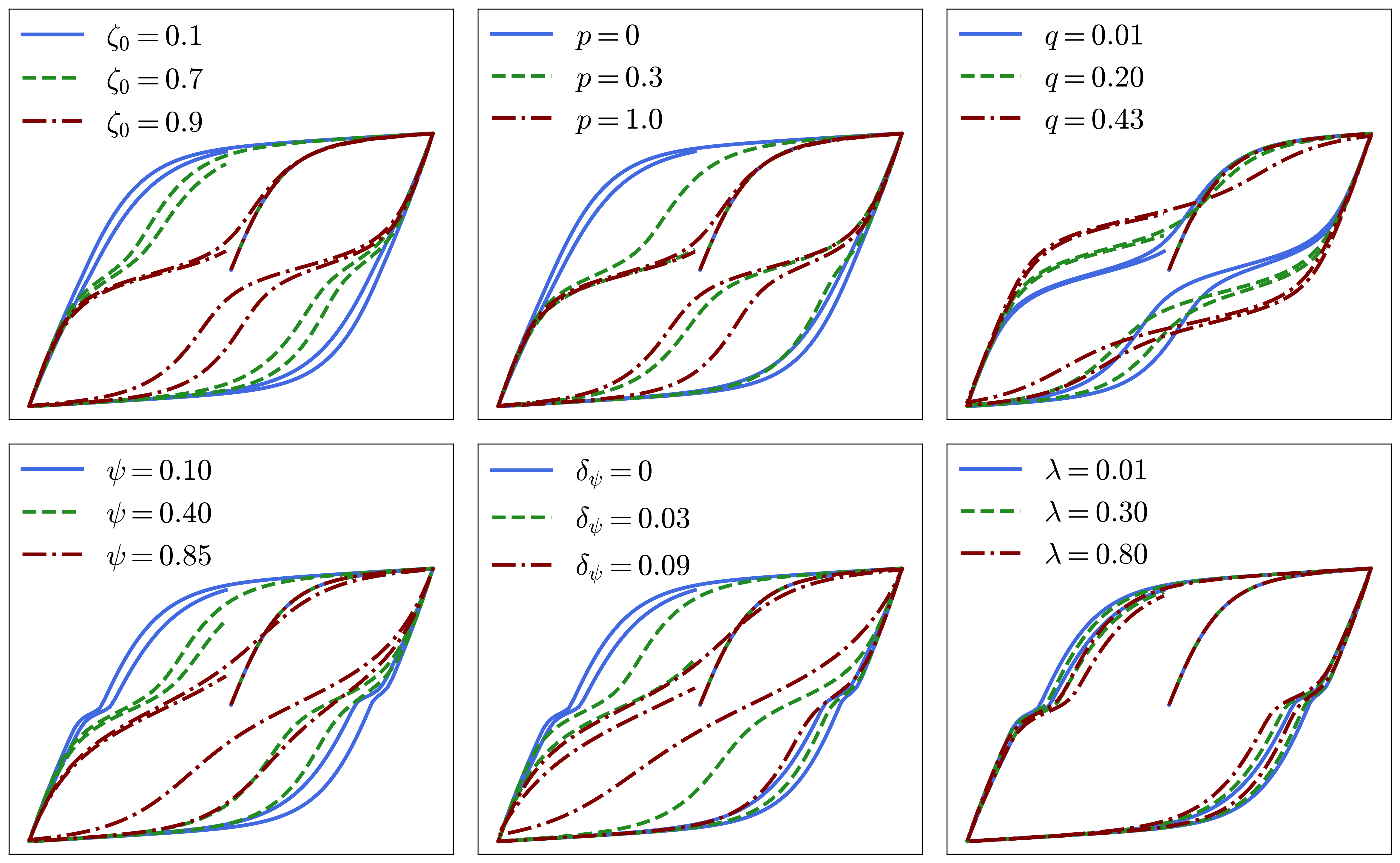}
        \caption{\textbf{Variations of the hysteresis curves for three different values of the PCH parameters.}}
        \label{fig:param_analy_pch}
    \end{figure}

    To illustrate the influence of individual parameters on the hysteresis curve, Figures \ref{fig:param_analy_bsc} to \ref{fig:param_analy_pch} show the variations of the curves for three different values of each parameter. While both BSC and DGD parameters primarily define the overall frame, such as its size, height, and narrowness, the PCH parameters specifically modulate changes in slope within the mid-range of the curve (i.e., around the origin). Because of this distinction in the influence domains between the parameter categories, we use the same CNN architecture for BSC and DGD, while a distinct architecture is employed for PCH. The following section provides details of the CNN architectures for each hysteresis category, along with their effectiveness as parameter estimators.

    \section{CNN architectures capturing hysteretic behaviors}\label{section:CNNmodels}
    \noindent    
    In this section describing the data generation process and CNN architectures, we refer to the structure of the neural network as the ``CNN architecture'' and the trained model from the architecture as the ``CNN model.'' The effectiveness of each CNN architecture is evaluated based on the estimation accuracy of the CNN model trained using the loading history provided in the guideline. Additionally, since the mean squared error, which is adopted as the loss function for training, provides limited insight into how well the predicted values align with the true values, the correlation coefficient between the true and predicted values is presented as a performance metric throughout this study, capturing the overall alignment. The training and test errors for each trained model are provided in \ref{appendix:mse_errors} to offer further details on the model performance.

    \subsection{Data generation}\label{section:data_generation}
    \noindent    
    A set of m-BWBN model parameters is generated using the probabilistic distributions summarized in Table \ref{table:params_mBWBN}. Based on the literature, uniform distributions are assumed for $T$, $F_y$, $\beta$, and $n$, while truncated normal distributions are adopted for $\alpha$, $\delta_\nu$, and $\delta_\eta$ to accommodate the empirical distributions derived from experimental results \cite{Zhang2022UncertaintyAnalysis, Kim2023DeepPinching, Zhang2024Hybrid-simulation-basedAnalysis}. A truncated joint normal distribution is employed for the PCH parameters to represent the complex interrelated relationship between them, where the distribution is constructed using the reinforced concrete column data in the Structural Performance Database \cite{Berry2004PEER1.2}. The histograms for the generated parameters are presented in \ref{appendix:histograms}, and the details of the distributions can be found in the shared code.
        
    A total of 500,000 data sets (m-BWBN model parameters and the corresponding hysteresis curves) are generated, with 450,000 used to train the CNN architectures and the remaining 50,000 for assessing estimation accuracy. When generating the training dataset, we intentionally add noise to address the overfitting issue. In other words, a random noise is added to each force value for a given displacement value. Four noise levels are introduced in terms of coefficients of variation: 0\% (no noise), 0.2\%, 0.5\%, and 0.8\%, representing zero, low, medium, and high noise levels, respectively. From the original 450,000 training data set, 300,000 are randomly selected for the zero, low, and medium noise levels, respectively, while 100,000 are selected for the high-noise hysteresis curves. This leads to a maximum of four hysteresis curves with different noise levels paired with the same parameter set. Moreover, a smaller number of high-noise data is used to avoid underfitting due to excessive noise. This results in a total of 1,000,000 training data sets.

    \subsection{CNN architectures}\label{section:cnn_architectures}
\begin{figure}[H]
        \centering
        \includegraphics[width=0.7\linewidth]{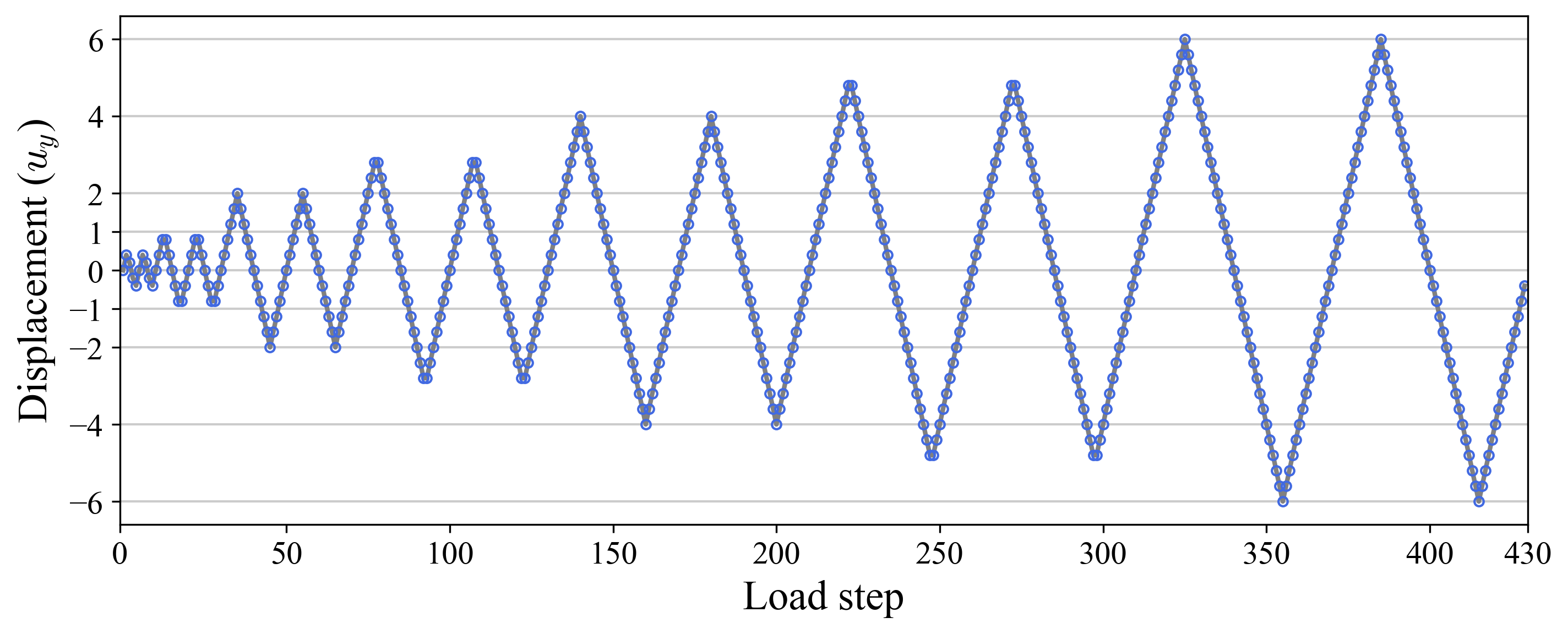}
        \vspace{-0.1in}
        \caption{\textbf{Loading history used for the CNN architecture development.}}
        \label{fig:CNN_architecture_loadinghistory}
    \end{figure}

    \noindent    
    The loading history sized $430\times1$ shown in Figure \ref{fig:CNN_architecture_loadinghistory}, normalized by the yield displacement $u_y$, is used to define the architecture. This loading history is adopted from the American Concrete Institute and Applied Technology Council \cite{ACICommittee3742013GuideLoads, ATC-241992GuidelinesStructures}, and is referred to as the reference loading history in this study. $u_y$ in Figure \ref{fig:CNN_architecture_loadinghistory} is defined as $\frac{F_y}{k_0}$ in the m-BWBN model. However, since there exist significant uncertainties in obtaining $u_y$ values in practice, we add noise with a coefficient of variation of 10\% to define $u_y$. For instance, for a parameter set with $F_y=0.5g$ and $T=0.3$ s, $u_y$ is determined as $\frac{0.5g}{\left(\frac{2\pi}{0.3}\right)^2} =1.12$ cm in the BW class model, and the final $u_y$ value is randomly sampled from a normal distribution with a mean of $1.12$ and a standard deviation of $0.112$.
    
    To effectively manage the different scales within the dataset during training of the CNN architectures, the hysteresis and m-BWBN parameters are  min-max normalized. A loss function is designed to minimize the difference between the true and the predicted parameter values.

    \subsubsection{Parameteres for basic hysteresis and degradation}\label{section:cnn_architectures_BSC_DGD}
\begin{figure}[H]
        \centering
        \includegraphics[width=0.9\linewidth]{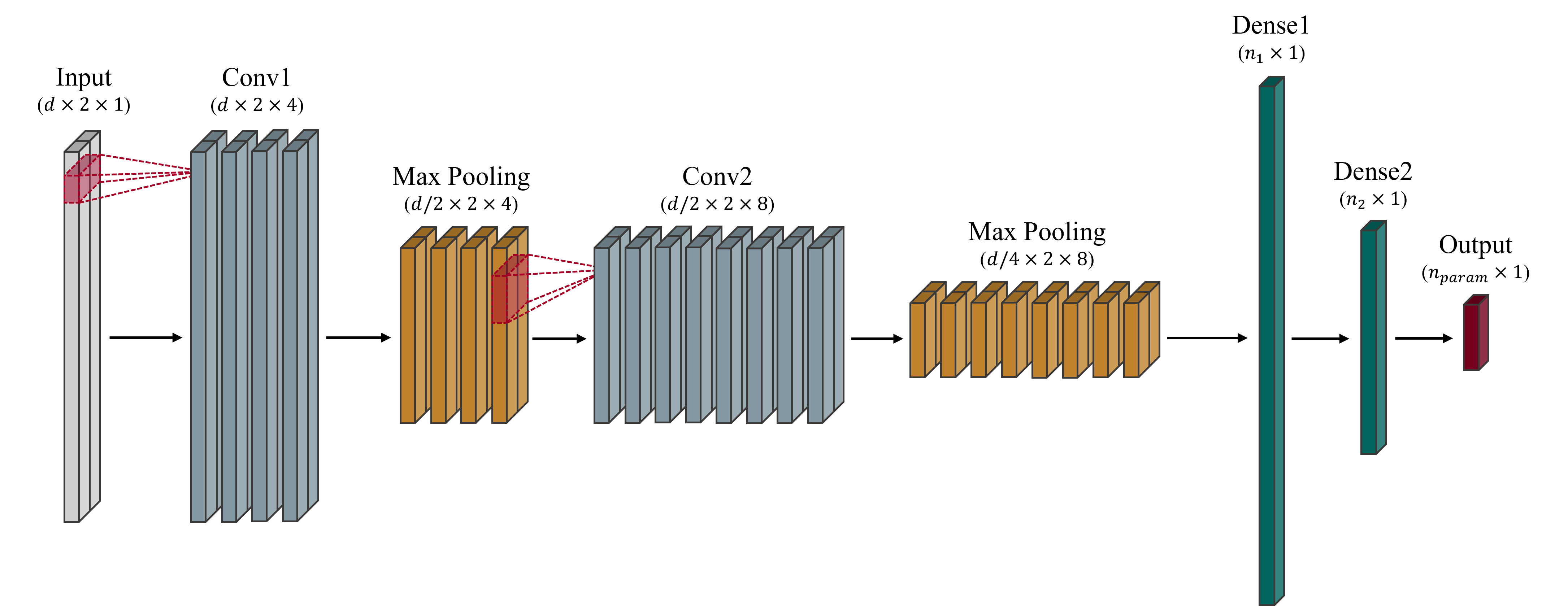}
        \caption{\textbf{CNN architecture for the BSC and DGD parameters.}}
        \label{fig:CNN_architecture_BSC_DGD}
    \end{figure}

    \noindent    
    Figure \ref{fig:CNN_architecture_BSC_DGD} outlines a CNN architecture for BSC and DGD parameters, where the input layer, sized at $d\times2\times1=430\times2\times1$, accommodates the 430 load steps from the loading history shown in Figure \ref{fig:CNN_architecture_loadinghistory}. In Figure \ref{fig:CNN_architecture_BSC_DGD}, $n_1$ and $n_2$ represent the number of nodes in the first and second dense layers, Dense1 and Dense2, respectively, which control the amount of transferred information from the convolution stages. $n_1=256$ and $n_2=32$ have been determined as an effective balance between model performance and complexity. The output dimension $n_{param}$ denotes the number of parameters, with $n_{param}=5$ for the BSC parameters and $n_{param}=2$ for the DGD parameters.

    The rectifier activation function (ReLU) activation function is used throughout the architecture, except in the output layer, where the sigmoid function is applied to constrain the outputs between 0 and 1. To maintain the input size during the convolution operations, the same padding command is applied. The convolutional layers are designed with 4 small $(2,2)$ and 8 larger $(4,2)$ filters, illustrated as red transparent boxes in Figure \ref{fig:CNN_architecture_BSC_DGD}, allowing the network to detect both fine and broader features within the hysteresis curve. Max pooling is performed after each convolutional layer to reduce the spatial dimensions while retaining key features identified by the convolution operations. A pool size of (2,1) is used to maintain the separation between displacement and resisting force information.
    
    The CNN architecture is trained separately for the BSC and DGD parameters, using the 1,000,000 training datasets. The Adam optimization algorithm is used with the mean squared error as the loss function. The epoch sizes of 100 and 300 are selected for the BSC parameters and the DGD parameters, respectively, with a batch size of 64 for both parameters.

\begin{figure}[H]
        \centering
        \begin{subfigure}{0.49\linewidth}
            \centering
            \includegraphics[height=6cm]{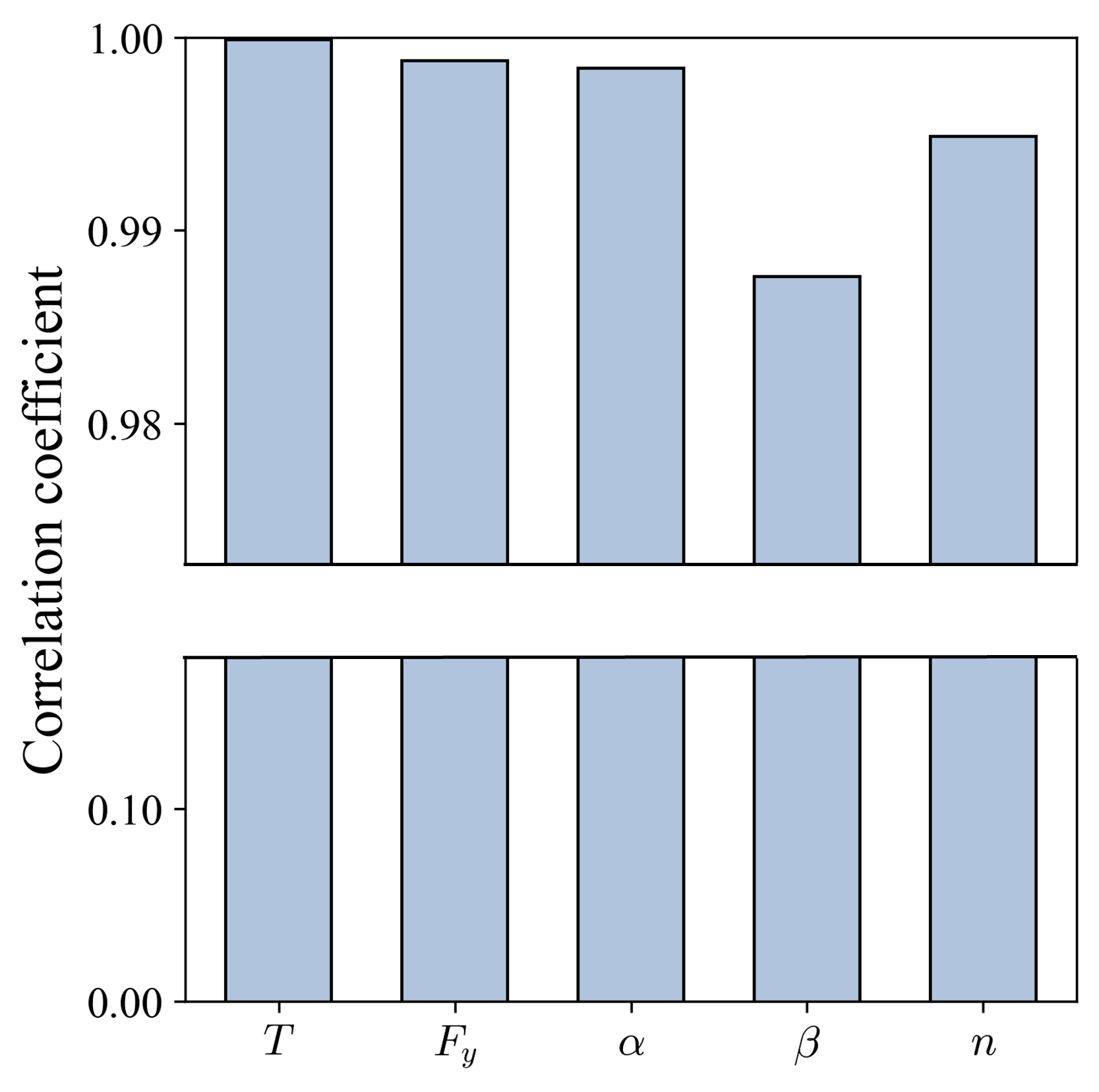}
            \vspace{-0.1cm}
            \caption{\textbf{}}
            \label{fig:barplot_BSC}
        \end{subfigure}
        \begin{subfigure}{0.49\linewidth}
            \centering
            \includegraphics[height=6cm]{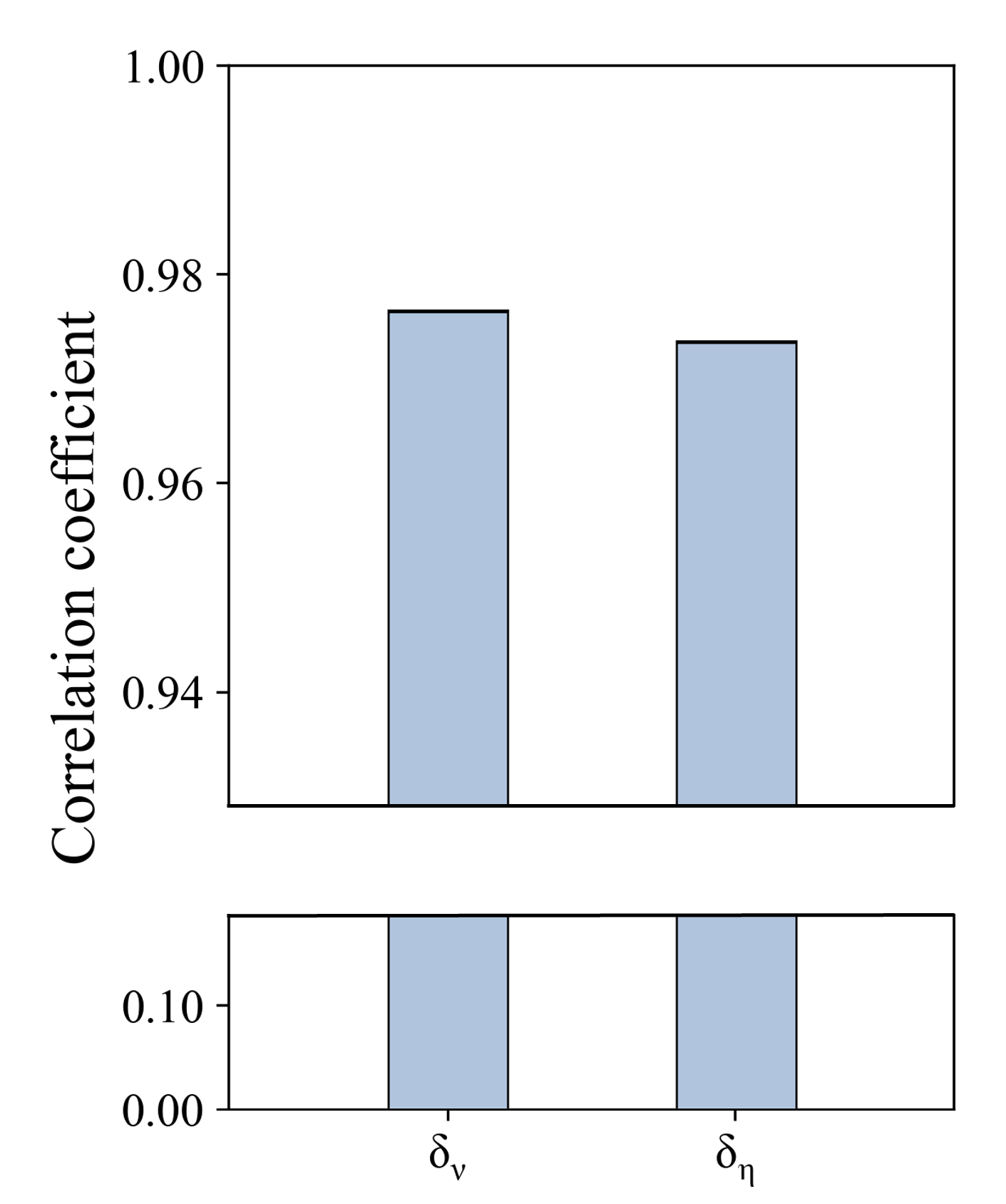}
            \vspace{-0.1cm}
            \caption{\textbf{}}
            \label{fig:barplot_DGD}
        \end{subfigure}
        \vspace{-0.1in}
        \caption{\textbf{Performance of the trained models demonstrated by correlation coefficients of (a) BSC parameters and (b) DGD parameters, respectively.}}
        \label{fig:barplot_BSC_DGD}
    \end{figure}

    The parameter estimation performance of the trained CNN models for BSC and DGD parameters is illustrated in Figures \ref{fig:barplot_BSC} and \ref{fig:barplot_DGD}, using the correlation coefficient values between the true and predicted values for the 50,000 test data. The results indicate significantly high correlation coefficients, all exceeding 0.95 for both BSC and DGD parameters. This strong predictive performance of the trained models highlights the effectiveness of the proposed CNN architecture in estimating the BSC and DGD parameters of a hysteresis curve.
    
    \subsubsection{Pinching parameters}\label{section:cnn_architectures_PCH}
\begin{figure}[H]
        \centering
        \includegraphics[width=0.9\linewidth]{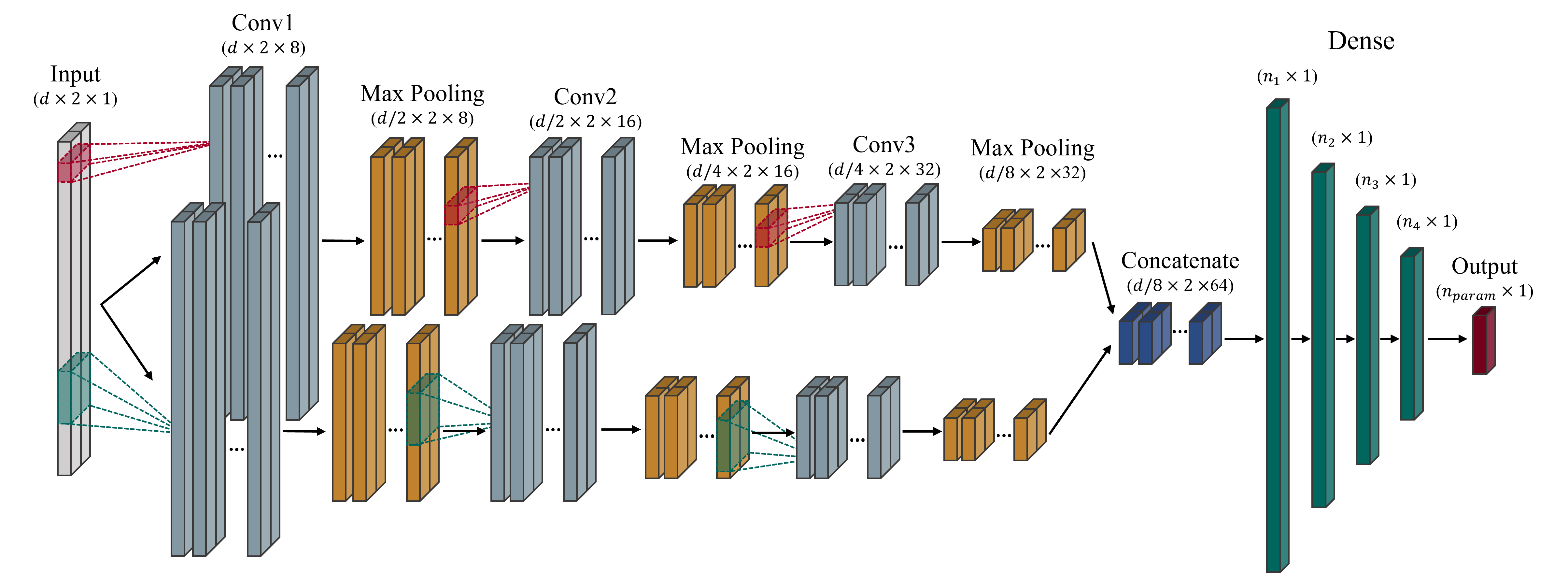}
        \caption{\textbf{The CNN architecture for the PCH parameters.} The input length $d$ depends on which loading history is applied. For the PCH parameters, $n_1=1024, n_2=256, n_3=64, n_4=16$ are used, and the number of parameters $n_{param}=6$.}
        \label{fig:CNN_architecture_PCH}
    \end{figure}

    \noindent
    The pinching effect modulates the stiffness, particularly in regions near zero displacement, and the PCH parameters primarily govern the spatial correlation between small- and large-displacement domains. The CNN architecture is designed to capture this spatial correlation as illustrated in Figure \ref{fig:CNN_architecture_PCH}.

    Compared to the previous architecture, two parallel convolutional operations, each with different filter sizes, are employed for the PCH parameters to capture the different scales of the two influence domains. The top convolution path, shown in Figure \ref{fig:CNN_architecture_PCH}, uses a filter size of $(2,2)$ throughout all layers, while the bottom path employs a filter size of $(16,2)$. In both paths, the number of filters doubles after each convolutional layer to minimize information loss, with Conv1, Conv2, and Conv3 using 8, 16, and 32 filters, respectively. The features extracted from these convolutional operations are then merged through a concatenation layer to predict the PCH parameters.

\begin{figure}[H]
        \centering
        \includegraphics[width=0.4\linewidth]{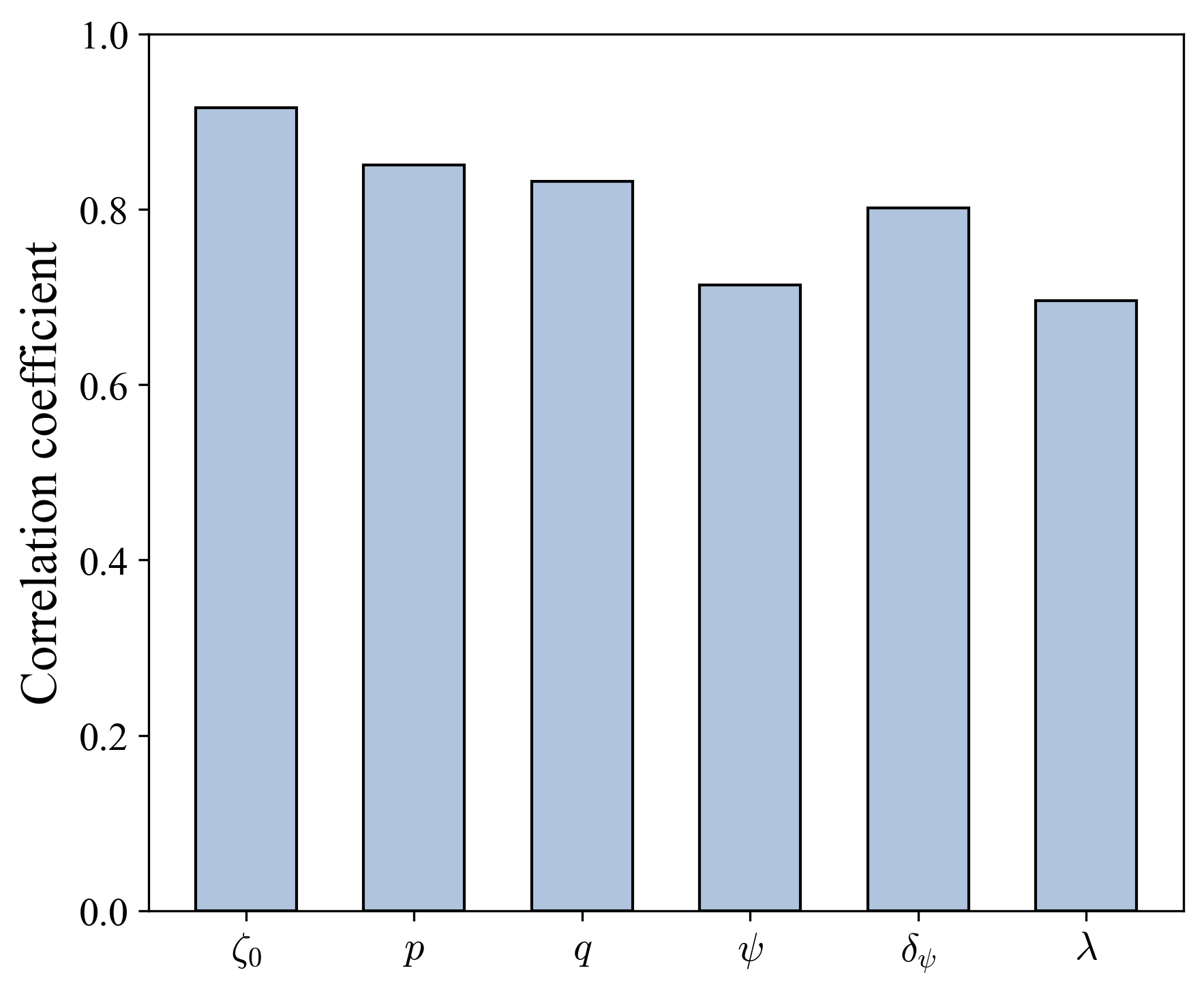}
        \vspace{-0.1in}
        \caption{\textbf{Performance of the trained models demonstrated by correlation coefficients of the PCH parameters.}}
        \label{fig:barplot_PCH} 
    \end{figure}

Similar to the BSC and DGD parameters, the CNN architecture of Figure \ref{fig:CNN_architecture_PCH} is trained with the 1,000,000 training data over 1,000 epochs with a batch size of 64. The same optimizer and the loss function used in the previous subsection are employed. The estimation performance of the trained CNN model for the PCH parameters is illustrated in Figure \ref{fig:barplot_PCH}.

    As shown in Figure \ref{fig:barplot_PCH}, $\zeta_0$ shows the highest correlation coefficient, followed by $p$, $q$, and $\delta_\psi$. However, the correlation coefficient values for $\psi$ and $\lambda$ are lower. This can be attributed to the fact that $\psi$ and $\lambda$ have the least influence on the pinching effect \cite{Kim2023DeepPinching, Li2021AModes, Yang2022ParameterColumns, Ajavakom2008PerformancePrediction}. In other words, even with certain uncertainties in these predicted values, the hysteresis curve predicted using the CNN architecture still aligns well with the true hysteresis. 
    
\begin{figure}[H]
    \centering
    \begin{subfigure}{0.95\textwidth}
        \centering
        \includegraphics[width=\textwidth]{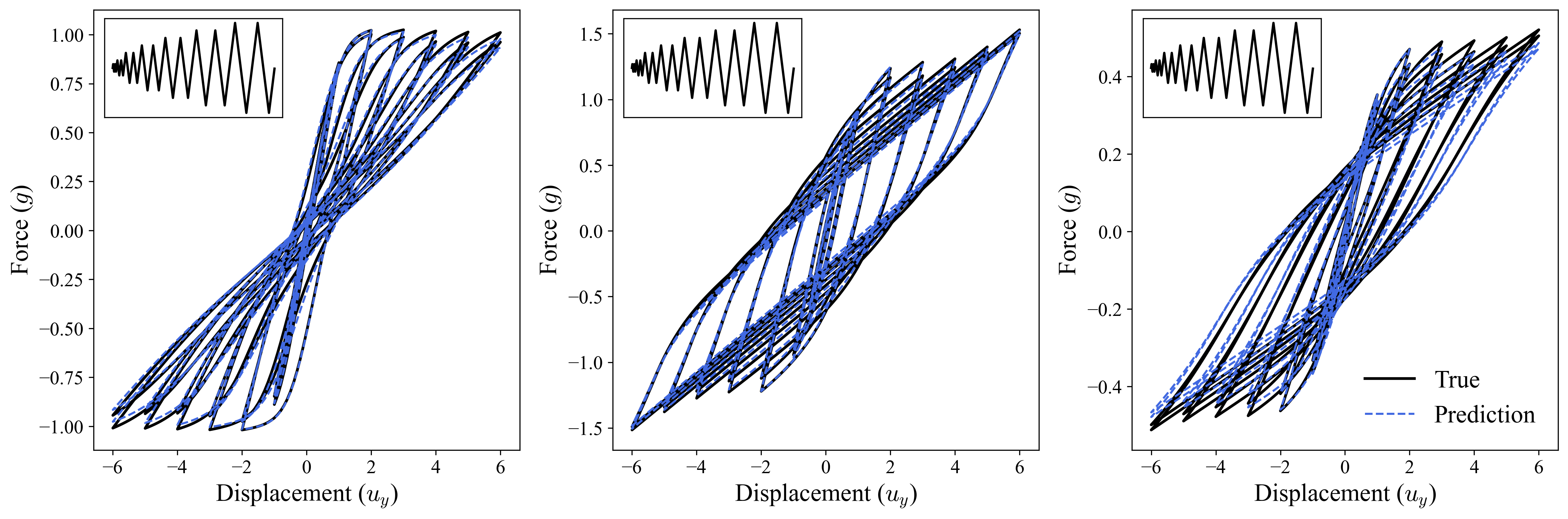}
        \caption{\textbf{}}
        \label{fig:total_hys_1}
    \end{subfigure}
    \begin{subfigure}{0.95\textwidth}
        \centering
        \includegraphics[width=\textwidth]{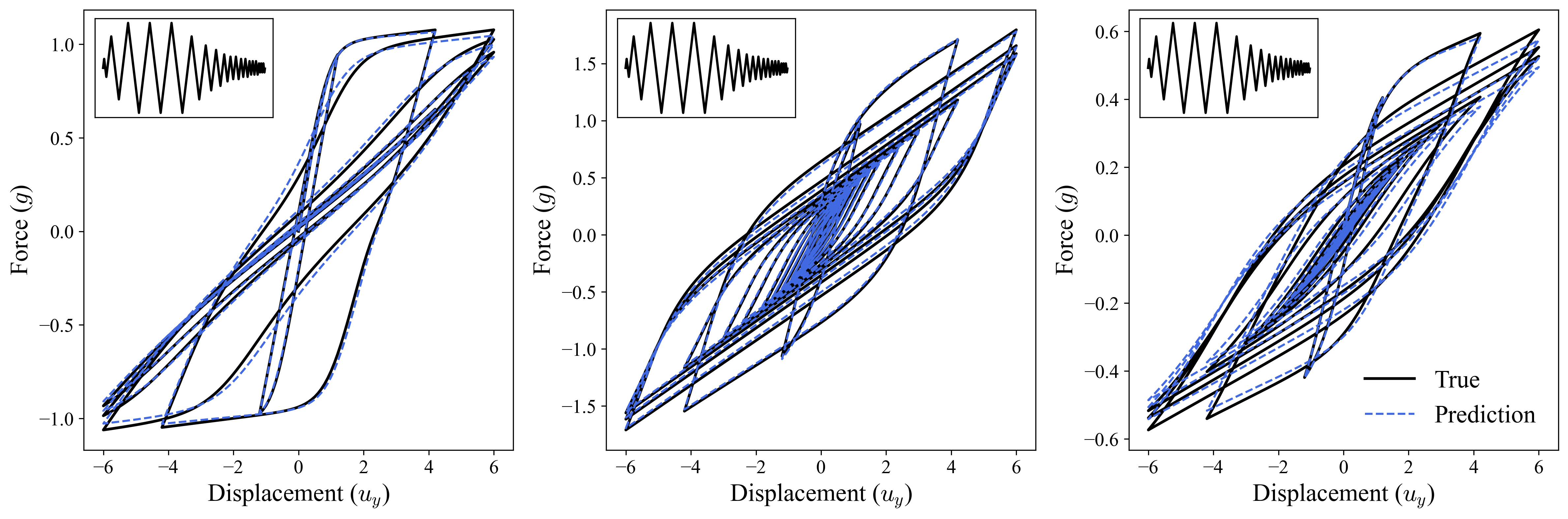}
        \caption{\textbf{}}
        \label{fig:total_hys_2}
    \end{subfigure}
    \caption{\textbf{Collective performance of trained CNN models: (a) the reference loading history and (b) the new loading history.}}
    \label{fig:total_hys}
    \end{figure}

    To evaluate the performance of the trained CNN models, Figure \ref{fig:total_hys} compares the true hysteresis curves with the predicted curves generated using the estimated parameters. Each BSC, DGD, and PCH parameter is estimated independently using the corresponding model and collected to form a complete set of parameters. The models are trained on the reference loading history. Three samples (represented as different columns in the figure) are randomly selected from the test dataset and applied to two different loading histories: the reference and a new loading histories, which are depicted in the upper left of each plot. The new loading history is the one follows an envelope function commonly used in synthetic ground motion analysis \cite{Jennings1968SimulatedMotions}. The true and predicted curves are represented by the black solid blue dashed lines, respectively. reliable performance of the trained CNN models shown in the six plots highlights the effectiveness of the proposed CNN architectures for the BSC, DGD, and PCH parameters.

\begin{figure}[H]
    \centering
    \begin{subfigure}{0.49\linewidth}
        \centering
        \includegraphics[height=6cm]{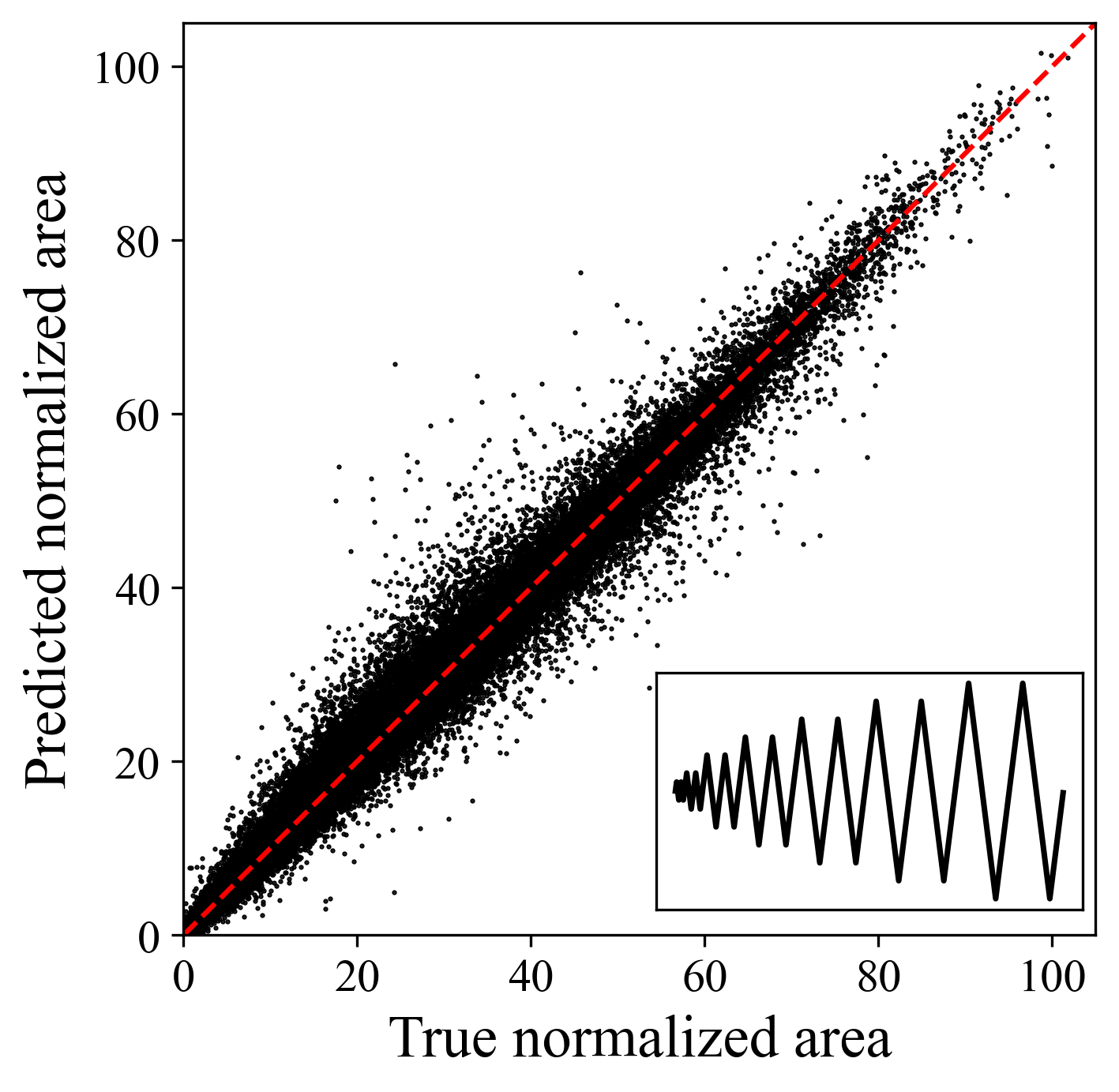}
        \caption{\textbf{}}
        \label{fig:total_scatter_1}
    \end{subfigure}
    \begin{subfigure}{0.49\linewidth}
        \centering
        \includegraphics[height=6cm]{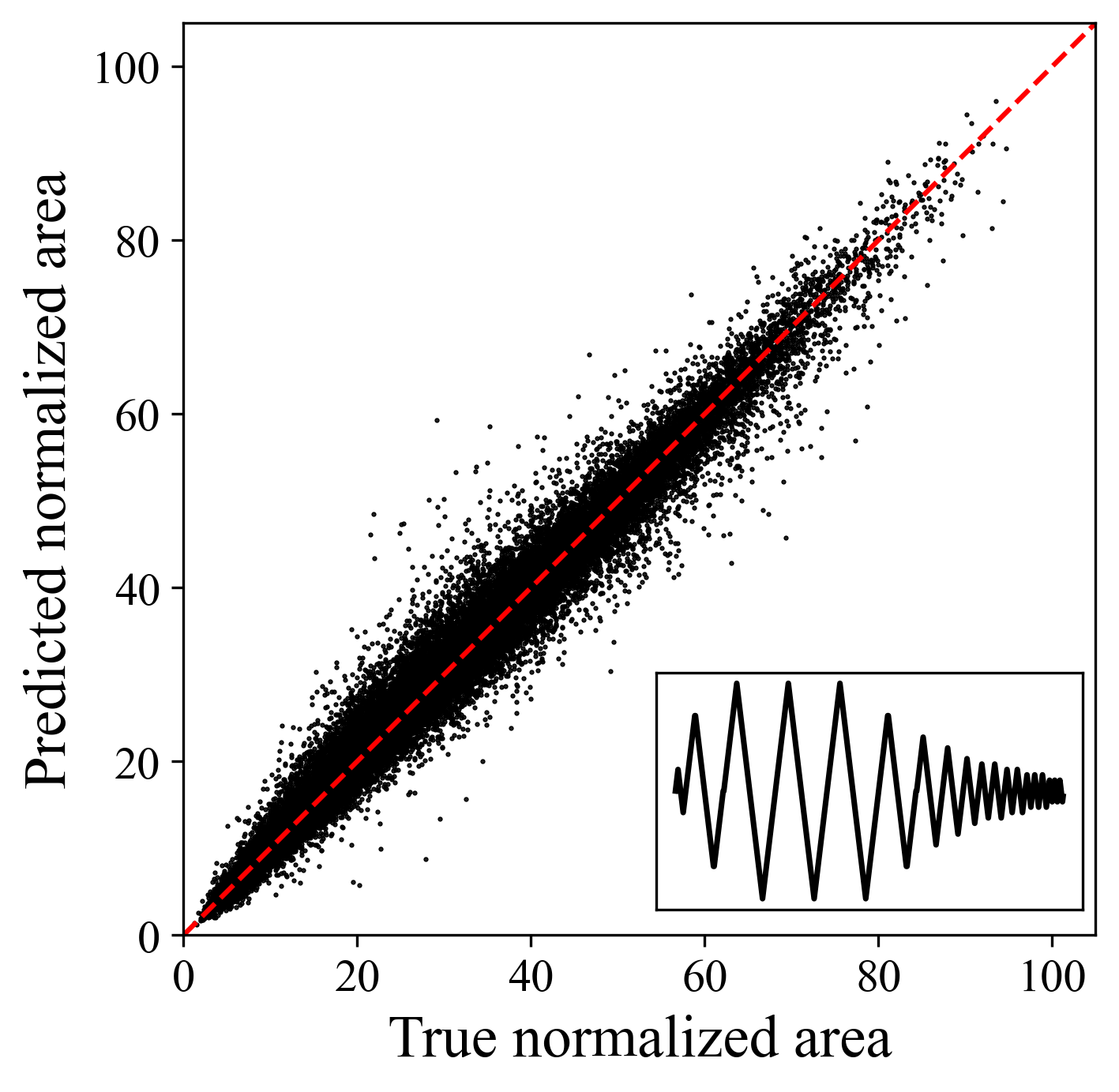}
        \caption{\textbf{}}
        \label{fig:total_scatter_2}
    \end{subfigure}
    \vspace{-0.1in}
    \caption{\textbf{Scatter plots of the true and predicted hysteresis energy for 50,000 parameter sets of the test data under two different loading histories: (a) the reference loading history and (b) the new loading history.}}
    \label{fig:total_scatter}
    \end{figure}

   To further investigate the performance of the CNN models, we present a scatter plot in Figure \ref{fig:total_scatter} comparing the true and predicted areas under the hysteresis curve for the entire test dataset, where the red dashed line indicates the identity line. The area is normalized by the yield strength $F_y$ and the yield displacement $u_y$ for consistent comparison across different scales of curves. The same two test loading histories as in Figure \ref{fig:total_hys} are employed, as depicted in the lower right of each plot. The results indicate that the area estimated using the predicted hysteresis closely matches the true value.

    \section{Loading history modules for different hysteresis categories} \label{section:loading_history}
    \noindent
    The CNN architectures presented in Section \ref{section:CNNmodels} are now trained on different loading histories to identify the loading history module for each hysteresis category. First, a set of loading histories is defined, and a distinct CNN model is constructed for each loading history. The relevance of these loading histories is evaluated based on the parameter estimation performance of the corresponding CNN model. After identifying the appropriate loading history module, optimal loading histories are proposed for the representative BW class models: the BW model, the BW model with degradation, and the BW model with degradation and pinching
    \subsection{Comprehensive loading history set}\label{section:loading_histories}
    \noindent
    Table \ref{table:loading_histories} summarizes the loading histories used to assess their impact on parameter estimation. A total of 18 loading histories are selected for comprehensive analysis, including the loading histories featuring a linear elastic range (LH1 and LH2), single cycles with varying amplitudes (LH3 to LH7), increasing cycle counts with constant amplitudes (LH8 to LH10), incrementally increasing amplitudes (LH11 to LH14), and constant cumulative displacement (LH15 to LH18), where the cumulative displacement refers to the sum of amplitudes over all cycles. The displacement per step is set at $0.1u_y$ for all loading histories.   
    \subsection{Basic hysteresis parameters}\label{section:loading_history_bsc}
\begin{figure}[H]
        \centering
        \includegraphics[width=\linewidth]{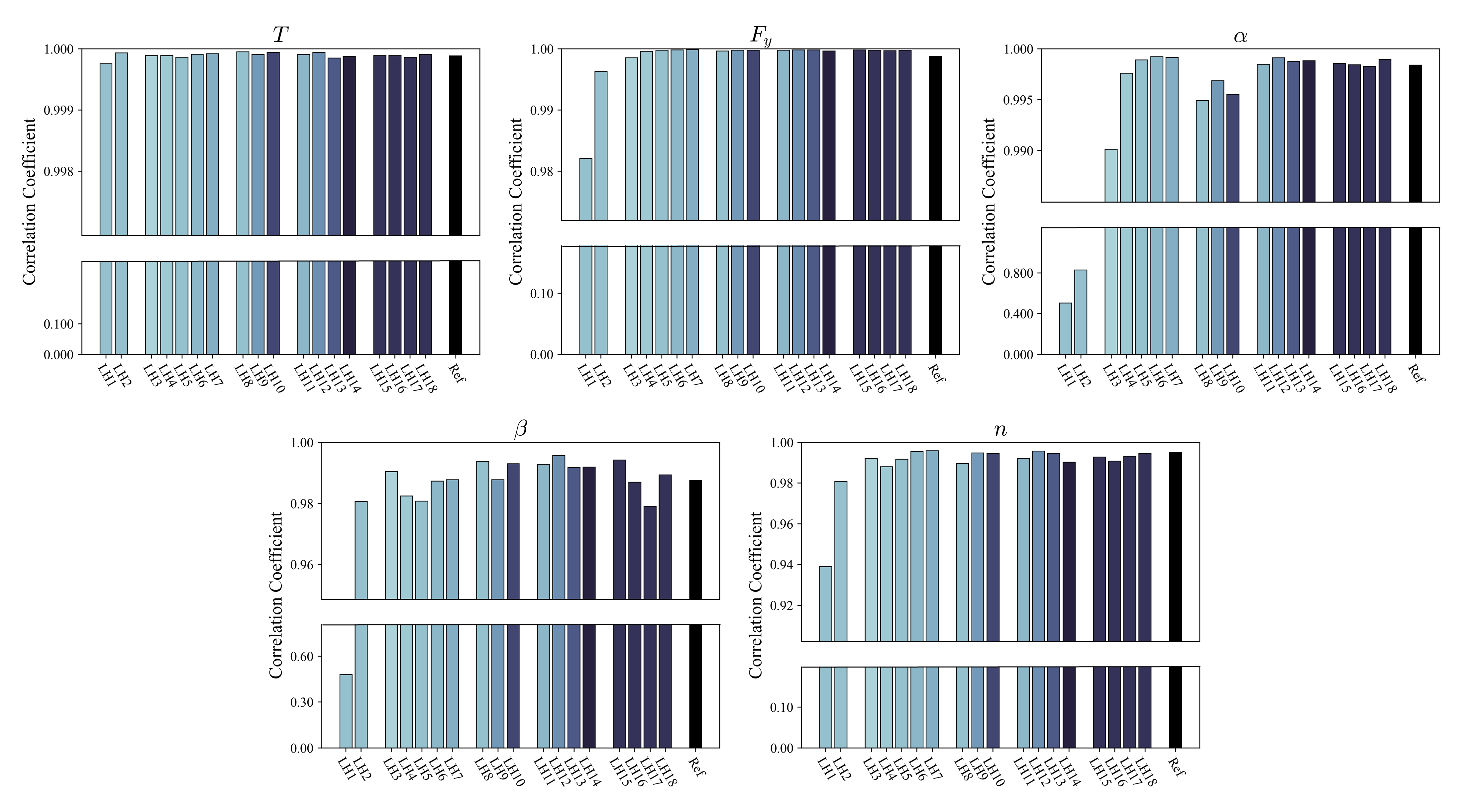}
        \vspace{-0.7cm}
        \caption{\textbf{Loading history analysis for the BSC parameters: correlation coefficients between the true and predicted values for 50,000 test data using the CNN models trained with different loading histories.}}
        \label{fig:LH_analy_bsc}
    \end{figure}

    \noindent
    After training the 18 CNN models on the loading history set, the parameter prediction performance of each loading history for the test dataset is presented in Figure \ref{fig:LH_analy_bsc} as a bar chart. The $y$-axis in the figure shows the correlation coefficient between the true and predicted parameter values. The bars are colored by cumulative displacement to account for the effect of the length of a loading history on prediction accuracy. While relatively small correlation coefficients are observed for loading histories with lower amplitudes, such as LH1 and LH2, except for $T$, most bars show significantly high values, indicating that the CNN model can accurately predict the BSC parameters. Specifically, loading histories confined to the linear elastic regions, LH1 and LH2, are sufficient to estimate the initial stiffness, represented as $T$. The yield strength $F_y$ similarly requires only a loading history that reaches or exceeds $u_y$ to be estimated with a correlation coefficient greater than 0.995. Notably, while $\beta$ and $n$ show no clear pattern after LH2, the post-to-pre-yield stiffness ratio $\alpha$ is better predicted with loading histories LH4-7 compared to LH8-10, despite the latter having larger cumulative displacements. This suggests that an amplitude of at least $3u_y$ is required for optimal estimation performance. In other words, to accurately estimate the BSC parameters, a loading history with an amplitude of $3u_y$ in at least one cycle is necessary, while two cycles are recommended as a conservative approach.
    
    \begin{table}[H]
    \caption{\textbf{Loading histories used to examine the impact of loading history on driving each hysteretic behavior.}}
    \label{table:loading_histories}
    \centering
    \begin{tabularx}{\linewidth}{@{}l>{\hsize=1.5\hsize}X>{\hsize=0.8\hsize}l>{\hsize=0.7\hsize}X@{}}
        \toprule[1.0pt]
        \textbf{Index} & \textbf{Amplitude sequence ($u_y$)} & \textbf{\begin{tabular}[c]{@{}l@{}}Cumulative\\displacement ($u_y$)\end{tabular}} & \textbf{Description} \\ \midrule[1.0pt]
        LH1 & $0.5-0.5-0.5-0.5-0.5-0.5-0.5-0.5$ & 4.0 & \multirow{2}{*}{\begin{tabular}[c]{@{}l@{}}Linear elastic range\end{tabular}} \\
        LH2 & $1.0-1.0-1.0-1.0$ & 4.0 &  \\
        \midrule
        LH3 & $2.0$ & 2.0 & \multirow{5}{*}{\begin{tabular}[c]{@{}l@{}}Single cycle\end{tabular}}  \\
        LH4 & $3.0$ & 3.0 &  \\  
        LH5 & $4.0$ & 4.0 & \\
        LH6 & $5.0$ & 5.0 &  \\
        LH7 & $6.0$ & 6.0 &  \\
        \midrule
        LH8 & $2.0-2.0$ & 4.0 &  \multirow{3}{*}{\begin{tabular}[c]{@{}l@{}}Increasing number of cycles\end{tabular}}  \\
        LH9 & $2.0-2.0-2.0-2.0$ & 8.0 &  \\
        LH10 & $2.0-2.0-2.0-2.0-2.0-2.0-2.0-2.0$ & 16.0 &  \\
        \midrule 
        LH11 & $2.0-3.0$ & 5.0 &  \multirow{4}{*}{\begin{tabular}[c]{@{}l@{}}Incrementally increasing amplitude\end{tabular}}  \\
        LH12 & $2.0-3.0-4.0$ & 9.0 &  \\ 
        LH13 & $2.0-3.0-4.0-5.0$ & 14.0 &  \\ 
        LH14 & $2.0-3.0-4.0-5.0-6.0$ & 20.0 &  \\
        \midrule
        LH15 & $2.0-2.0-3.0-3.0-4.0-4.0$ & 18.0 &  \multirow{4}{*}{\begin{tabular}[c]{@{}l@{}}Constant cumulative displacement\end{tabular}}  \\
        LH16 & $2.0-2.0-2.0-4.0-4.0-4.0$ & 18.0 &  \\ 
        LH17 & $2.0-2.0-2.0-3.0-4.0-5.0$ & 18.0 &  \\ 
        LH18 & $2.0-2.0-2.0-2.0-5.0-5.0$ & 18.0 &  \\
        \bottomrule[1.0pt]
    \end{tabularx}
    \end{table}

    \subsection{Degradation parameters}\label{section:loading_history_dgd}
\begin{figure}[H]
        \centering
        \includegraphics[width=0.75\linewidth]{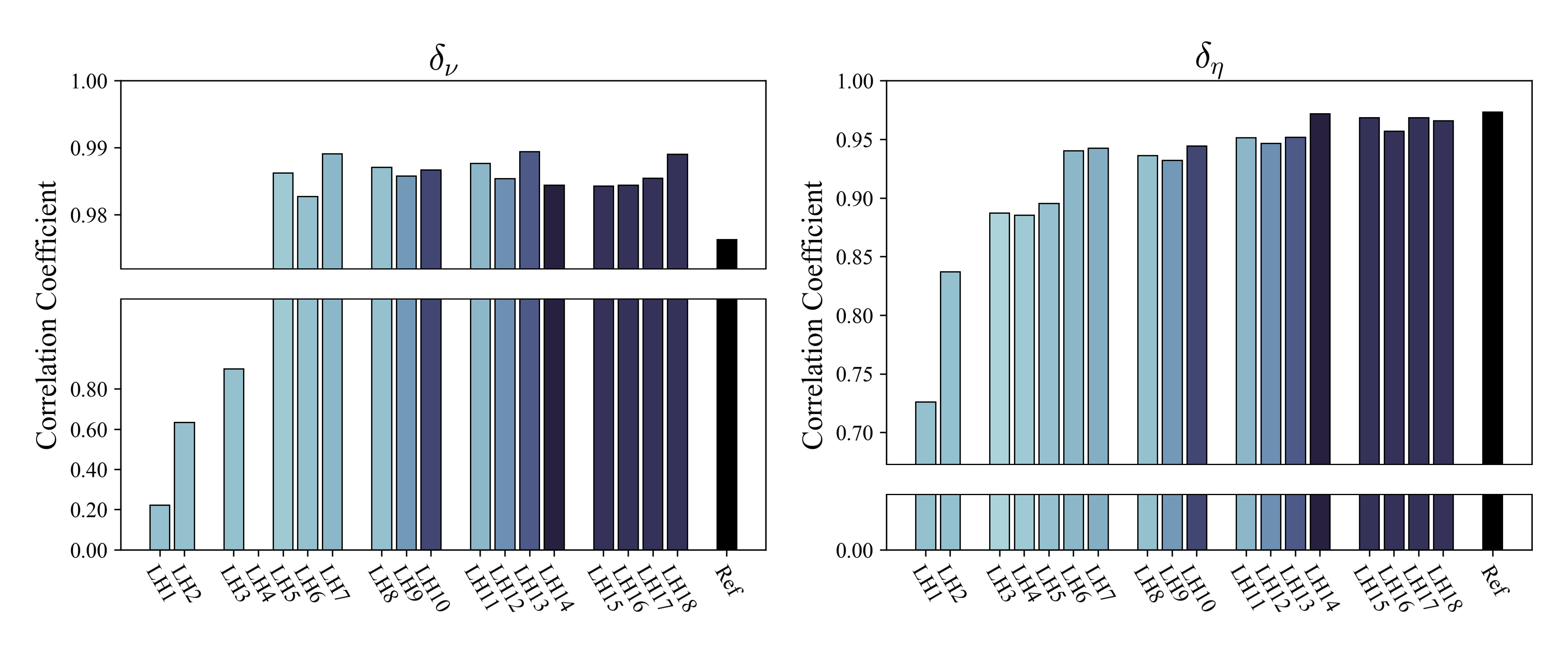}
        \vspace{-0.3cm}
        \caption{\textbf{Loading history analysis for the DGD parameters: correlation coefficients between the true and predicted values for 50,000 test data using the CNN models trained with different loading histories.}}
        \label{fig:LH_analy_dgd}
    \end{figure}

    \noindent
    The bar plots in Figure \ref{fig:LH_analy_dgd} illustrate the DGD parameter estimation results. Since LH1 and LH2 do not have a loading history exceeding the yield strength (i.e., confined to the elastic range), they show limited performance in predicting the DGD parameters. 
     
    In the BW class model, structural degradation intensifies as the loading history continues, as described in Eqs. \eqref{eq:epsilon_n} to \eqref{eq:nu}. However, from the perspective of CNN model training, having a longer loading history (i.e., more displacement steps) does not necessarily lead to better prediction performance, especially when the added information is irrelevant to the output. This trend can be further amplified if the input contains noise, as in this study, which explains the unclear increasing trend in $\delta_\nu$ with cumulative displacement. Finally, based on the observation that 1) the loading histories with cumulative displacements greater than $5u_y$ yield reliable correlation coefficient values that are larger or comparable to those of the reference loading history, and 2) the darker bars show overall taller heights for $\delta_{\eta}$, a loading history with a cumulative displacement of $10u_y$ is identified as a loading history module for DGD parameters, as a conservative approach.
   
    Moreover, the correlation coefficient values for the strength degradation rate $\delta_\nu$ are generally larger than those for the stiffness degradation rate $\delta_\eta$. This stems from the more entangled mechanism of stiffness degradation in the m-BWBN model, as shown in Eqs.\eqref{eq:z_ode}, \eqref{eq:eta}, and \eqref{eq:nu}, where the effect of $\delta_\eta$ is intertwined with the pinching function, while $\delta_\nu$ directly contributes to the maximum height of $z$. 

    \subsection{Pinching parameters}\label{section:loading_history_pch}
\begin{figure}[H]
        \centering
        \includegraphics[width=1.0\linewidth]{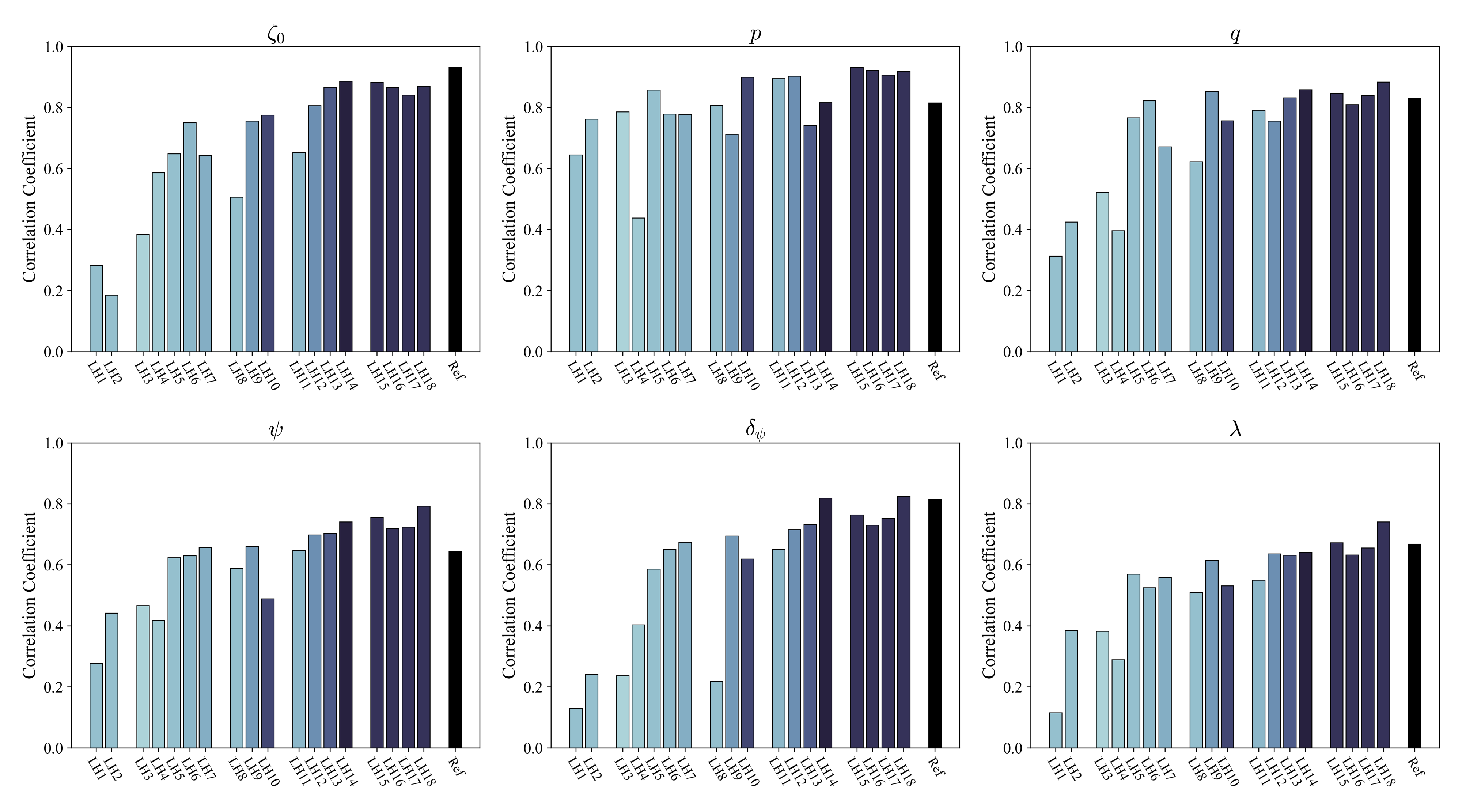}
        \vspace{-0.7cm}
        \caption{\textbf{Loading history analysis for the PCH parameters: correlation coefficients between the true and predicted values for 50,000 test data using the CNN models trained with different loading histories.}}
        \label{fig:LH_analy_pch}
    \end{figure}

    \noindent
    Figure \ref{fig:LH_analy_pch} presents the performance of trained CNN models to predict the PCH parameters under various loading histories. It is observed that the darker bars, which represent larger cumulative displacements, generally predict the parameters with higher correlation coefficients. This is consistent with the mathematical mechanism of the pinching effect in the m-BWBN model, as described Eqs.\eqref{eq:pinching} to \eqref{eq:zeta2}, where the pinching effect is governed by cumulative hysteretic energy $\varepsilon_n$ which generally increases as loading continues. Note that, similar to Figure \ref{fig:barplot_PCH}, the overall correlation coefficient values in Figure \ref{fig:LH_analy_pch} are lower than those for the BSC and DGD parameters. However, as discussed earlier, the CNN model’s performance in predicting the PCH parameters remains sufficient for capturing the overall hysteresis curve.
    
    Based on the numerical investigation, it is found that LH14 with a cumulative displacement of $20u_y$ does not show significant performance improvements compared to LH15-18 which has cumulative displacements of $18u_y$. Thus, we conclude that a loading history with a cumulative displacement of $18u_y$ is sufficient to predict the pinching effect in the m-BWBN model. Given that loading histories with gradually increasing amplitudes are commonly used in parameter estimation, LH15 is suggested as the loading history module for PCH parameters.

    \subsection{Proposed loading protocol}\label{section:proposed_loading_protocol}
\begin{figure}[H]
        \centering
        \includegraphics[width=0.9\linewidth]{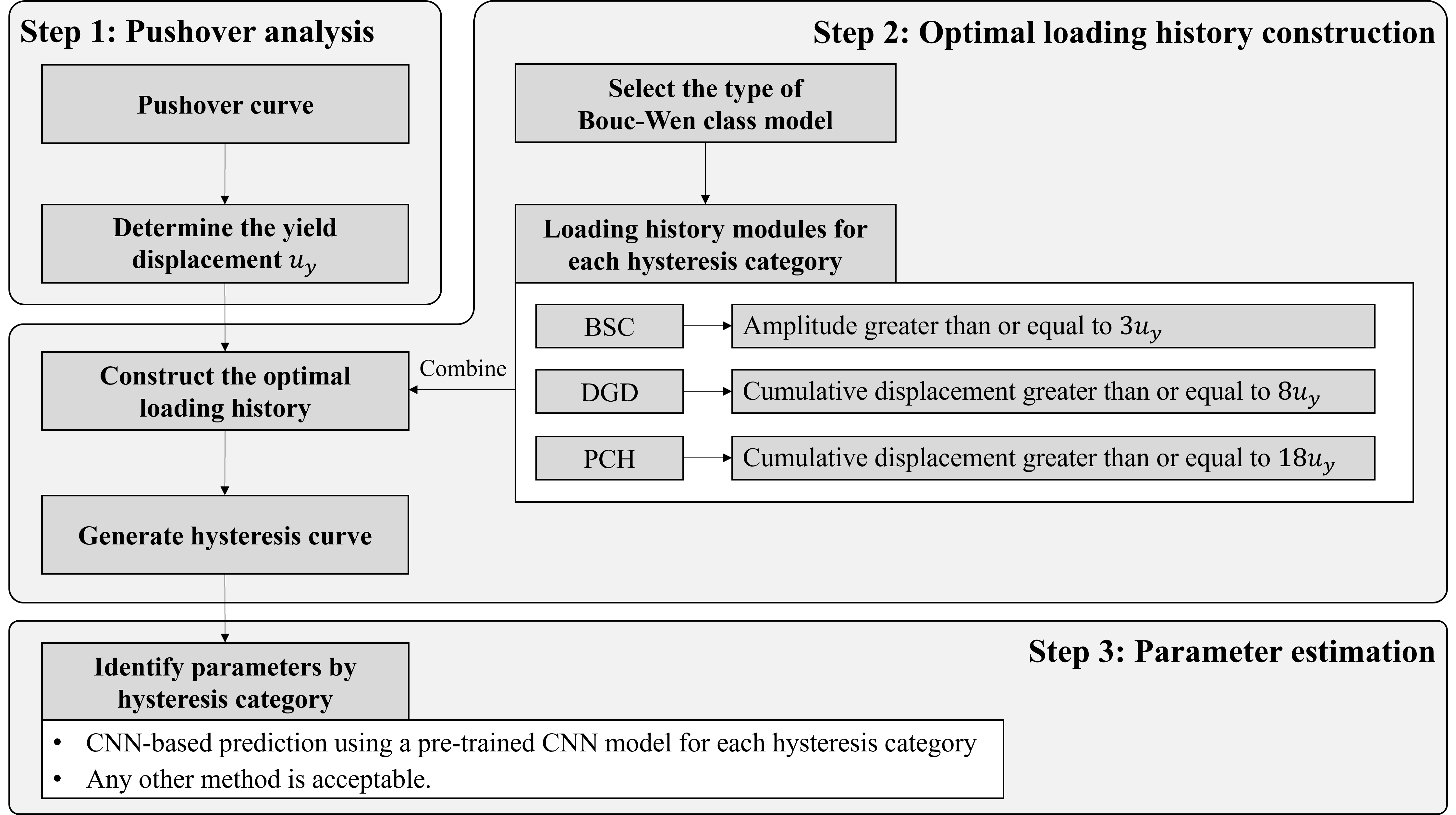}
        \caption{\textbf{Proposed modularized loading protocol.}}
        \label{fig:proposed_loading_protocol}
    \end{figure}

    \noindent
    Figure \ref{fig:proposed_loading_protocol} presents the modularized loading protocol based on the loading history modules for three different hysteresis categories. The protocol consists of three steps: 1) pushover analysis to determine the yield displacement, 2) construction of the optimal loading history, and 3) parameter estimation. Since the loading history modules identified in Sections \ref{section:loading_history_bsc} to \ref{section:loading_history_pch} are defined in terms of $u_y$, it is necessary to define $u_y$. While several approaches exist \cite{FEMA3562000PRESTANDARDBUILDINGS, DeLuca2013Near-optimalAnalysis}, this study adopts a method based on the pushover curve. In particular, $u_y$ is determined as the intersection point between two lines: one is the linear extension from the origin with a slope of initial stiffness, and the other is a line with the lowest positive tangential slope along the pushover cover. The visual illustration of the process for determining $u_y$ is provided in the numerical investigation section, especially using Figures \ref{fig:Example1_pushover} and \ref{fig:Example2_pushover}. 

    Once $u_y$ is determined, the optimal loading history is constructed by combining the relevant loading history modules. Only the modules corresponding to the hysteretic characteristics of the target hysteresis model are employed, facilitating this step modular and adaptable to different BW class models. Lastly, the parameters are estimated from a hysteresis curve obtained under the constructed optimal loading history. While the CNN models trained on the loading history module serve as rapid parameter estimators, the proposed protocol offers flexibility, allowing the use of other methods, including genetic algorithm-based estimation.

\begin{figure}[H]
        \centering
        \includegraphics[width=0.6\linewidth]{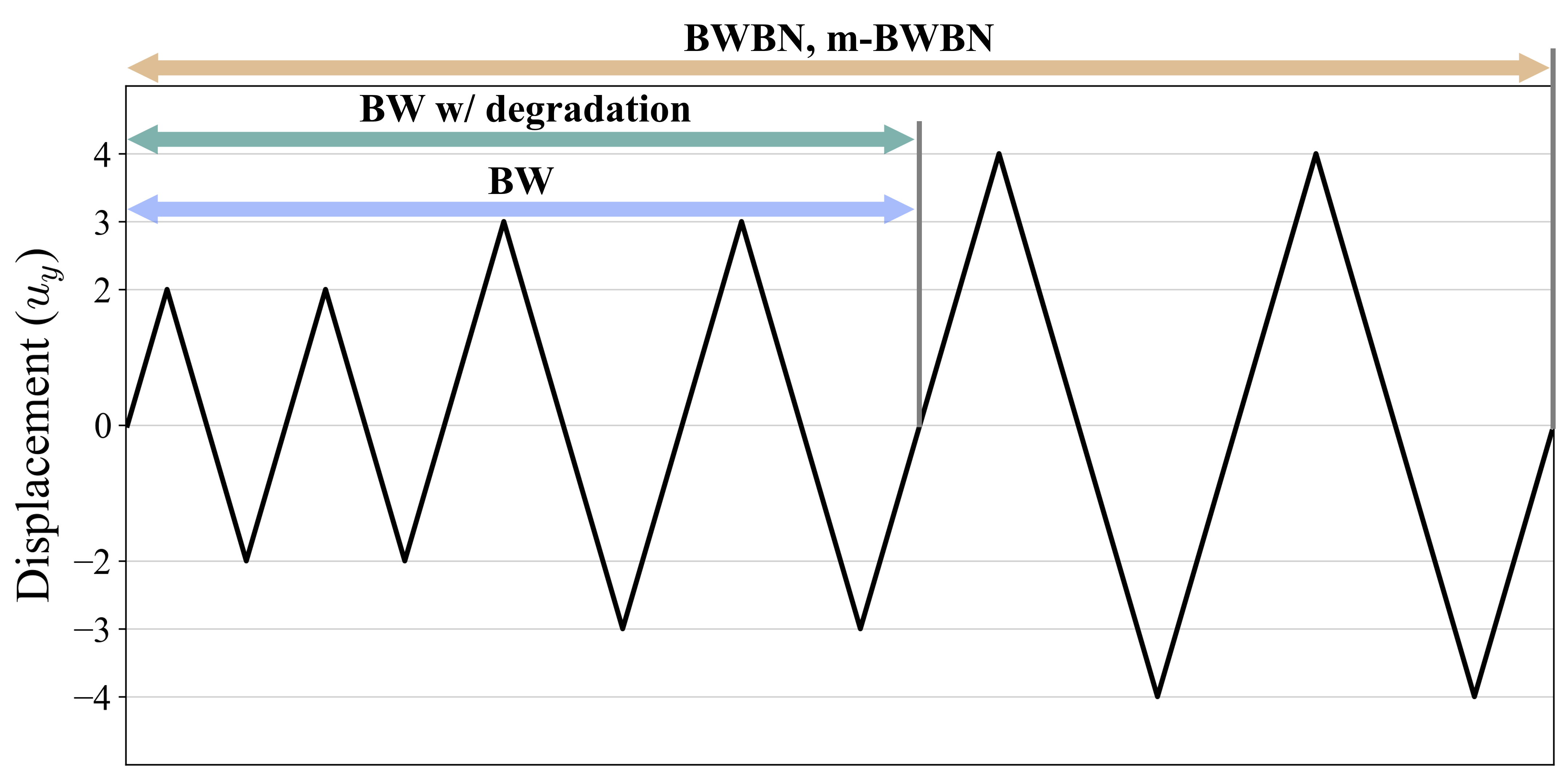}
        \caption{\textbf{Examples of optimal loading histories for different Bouc-Wen class models.}}
        \label{fig:optimal_loading_history}
    \end{figure}

    Furthermore, to facilitate the practical use of the protocol, we propose optimal loading histories for four different BW class models: (1) BW, (2) BW with degradation, (3) BWBN, and (4) m-BWBN models. Note that the BW model consists of only the BSC parameters, while the BW model with degradation includes the BSC and DGD parameters. The BWBN and m-BWBN models involve all three hysteresis categories. These displacement histories are illustrated in Figure \ref{fig:optimal_loading_history}. 
    
    The optimal loading history for the BW model, shown in Figure \ref{fig:optimal_loading_history}, consists of two cycles at $2u_y$ followed by two cycles at $3u_y$, satisfying the recommendation for reliable BSC parameter estimation outlined in Section \ref{section:loading_history_bsc}. For the BW model with degradation, commonly used to represent steel structures and base isolators \cite{Ismail2009TheSurvey, Sengupta2017HysteresisArt}, the loading history should incorporate the characteristics required for both BSC and DGD parameters, as demonstrated in Figure \ref{fig:optimal_loading_history}, which features a cumulative displacement of $10u_y$. Additionally, the optimal loading history for BW class models involving both degradation and pinching effect is depicted in Figure \ref{fig:optimal_loading_history}, addressing all the conditions for each hysteresis category. Because the optimal loading histories for BSC and DGD parameters are a subset of that for PCH parameters, only a single hysteresis curve is required to estimate parameters from all three hysteresis categories.

    Note that modifications to the proposed loading histories are needed if hysteretic characteristics not considered in this study, such as acute deterioration and pinching relaxation in reinforced concrete columns \cite{Oh2023BoucWenAnalysis}, are considered. To identify the characteristics of a loading history that induces the desired hysteresis behavior and to construct the corresponding optimal loading history, the same procedures used in this study from Sections \ref{section:CNNmodels} to \ref{section:loading_history} can be repeated.

    \section{Numerical investigations}\label{section:numerical_investigations}
    \noindent
    Three numerical investigations are performed to evaluate the efficiency and effectiveness of the proposed protocol. First, the effectiveness of the optimal loading history is demonstrated by comparing the parameter estimation accuracy obtained from the optimal and reference loading histories using a genetic algorithm. Subsequently, the parameter estimation performance of the CNN models with the optimal loading history is assessed through two structural systems: a 3-story SAC structure and a 3-story reinforced concrete (RC) frame structure. These two structures are chosen to assess the effectiveness of the proposed framework for common structural types, in particular a steel structure and an RC structure. The computation times reported in this section are based on a 13th Gen Intel\textsuperscript{\textregistered} Core\textsuperscript{TM} i9-13900 @ 2.50GHz.

    \subsection{Optimal loading history demonstration using genetic algorithm}\label{section:opt_loading_history_demo}
    \noindent  
    To demonstrate the effectiveness of the optimal loading histories, the following three different BW class models are introduced: (1) BW model, (2) BW model with degradation, and (3) m-BWBN model. For each model, 1,000 hysteresis curves are generated using both the optimal loading history suggested in Figure \ref{fig:optimal_loading_history} and the reference loading history shown in Figure \ref{fig:CNN_architecture_loadinghistory}. The parameters of each model are estimated from these 1,000 hysteresis curves and then used to predict the hysteresis curve for a new loading history for validation. The loading history following the envelope function, used in Figures \ref{fig:total_hys_2}, is adopted as the new loading history for testing.

    The comparison between the true and predicted hysteresis curves is summarized in Table \ref{table:LH_demo_summary}. The accuracy is defined as the percentage of cases where the predicted normalized area under the hysteresis curve falls within a specified error range (5\% and 10\%) out of 1,000 data, where the area is normalized by the yield strength $F_y$ and the yield displacement $u_y$. The scatter plots showing detailed results are provided in \ref{appendix:scatter_plots}. The results show that the parameter estimation using the optimal loading history yields accuracy values comparable to those achieved with the reference loading history that contains more displacement steps and higher amplitudes. Notably, for the m-BWBN model, the optimal loading history surpasses the reference in estimation accuracy. The effectiveness of the optimal loading history is further highlighted when the elapsed time to estimate the parameters is taken into consideration: due to its redundant length and information, the reference loading history requires 3.5 to 6 times longer time to estimate parameters. This time reduction can be more pronounced when parameter estimation is needed for a large number of structures, such as in a regional-scale seismic simulation \cite{Zhong2024RegionalPortfolios, Xu2023Regional-scaleNetwork, Steelman2009InfluenceEstimation, McKenna2024NHERI-SimCenter/R2DTool:4.0.0}.

    \begin{table}[H]
    \caption{\textbf{Summary of the optimal loading history demonstration results. BWdeg represents the BW model with degradation.}}
    \label{table:LH_demo_summary}
    \centering
        \begin{tabular}{llccc}
        \toprule[1pt]
        \multirow{2}{*}[-0.6ex]{\textbf{Model}} & \multirow{2}{*}[-0.6ex]{\textbf{Loading history}}    & \multicolumn{2}{c}{\textbf{Accuracy}}                  & \multirow{2}{*}[-0.6ex]{\textbf{Mean elapsed time}}   \\ \cmidrule{3-4}
                                                &                                                      & \textbf{Within 10\% error}& \textbf{Within 5\% error}  &                                                       \\
        \midrule[1pt]
        \multirow{2}{*}{BW}                     & Optimal                                              &  99.3\%                   &  91.7\%                    &  461.49 s                                             \\
                                                & Reference                                            &  100.0\%                  &  99.7\%                    &  1861.37 s                                            \\
        \midrule
        \multirow{2}{*}{BWdeg}                  & Optimal                                              &  96.5\%                   &  78.2\%                    &  4903.22 s                                            \\
                                                & Reference                                            &  98.3\%                   &  81.4\%                    &  18344.98 s                                           \\
        \midrule
        \multirow{2}{*}{m-BWBN}                 & Optimal                                              &  92.5\%                   &  71.5\%                    &  13755.36 s                                          \\
                                                & Reference                                            &  92.3\%                   &  70.6\%                    &  51367.31 s                                           \\
        \bottomrule[1pt]
        \end{tabular}
    \end{table}    \subsection{3-story SAC structure}
    \noindent
    A 3-story steel moment frame building specified by the 1994 SAC Steel Project, referred to as the 3-story SAC structure, is introduced \cite{Gupta1999SEISMICSTRUCTURES, Goel2004EvaluationBuildings, Deierlein1998SUMMARYANALYSES, Zhou2022Energy-basedDampers}. OpenSees is employed to model the 3-story SAC structure \cite{McKenna2011OpenSees:Simulation}.

\begin{figure}[H]
        \centering
        \includegraphics[height=6cm]{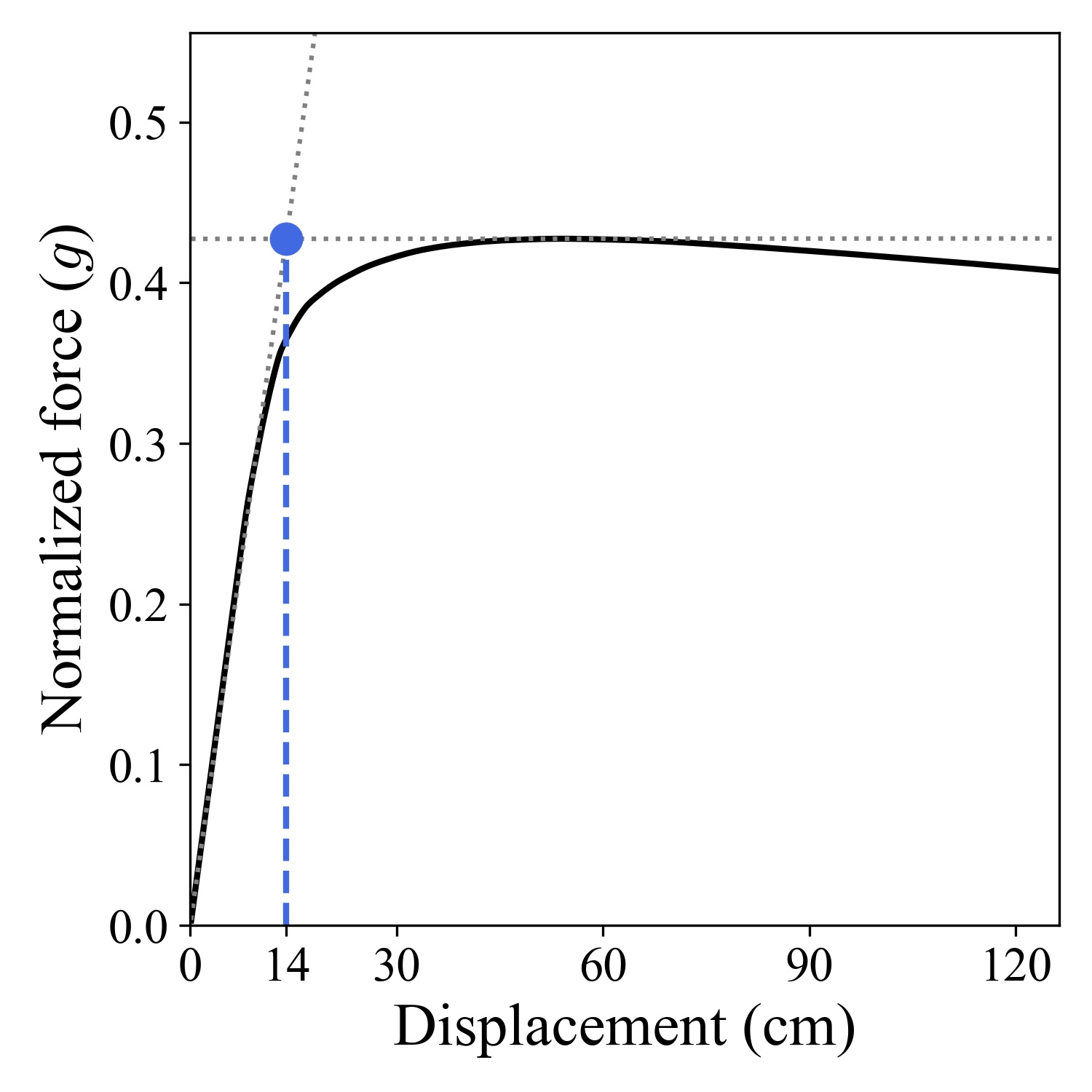}
        \vspace{-0.1in}
        \caption{\textbf{A pushover curve of the base shear force and roof displacement for the 3-story SAC structure.}}
        \label{fig:Example1_pushover}
    \end{figure}

    Figure \ref{fig:Example1_pushover} illustrates the pushover curve for the 3-story SAC structure, where the base shear force is normalized by the mass of the structure. The yield displacement $u_y$ is estimated to be 14 cm following the method outlined in Section \ref{section:proposed_loading_protocol}. 

    Since structural degradation and the pinching effect are not considered in the OpenSees model, the BW model is adopted to represent the force-displacement relationship between the base shear and roof displacement. Accordingly, the optimal loading history for the BW model is used. The parameters of the BW model are estimated using both a genetic algorithm and the trained CNN model (only corresponding to the BSC parameters) for comparison. In the genetic algorithm, the number of iterations (generations) is set to 100, with each generation consisting of 300 samples (populations), and the mean squared error between the predicted and actual force given the displacement history is used as the fitness function.

\begin{figure}[H]
        \centering
        \begin{subfigure}{0.49\linewidth}
            \centering
            \includegraphics[height=5cm]{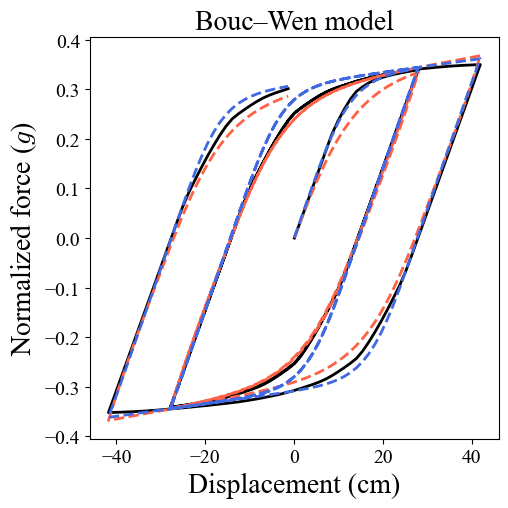}
            \caption{\textbf{}}
            \label{fig:Example1_hys_BW}
        \end{subfigure}
        \begin{subfigure}{0.49\linewidth}
            \centering
            \includegraphics[height=5cm]{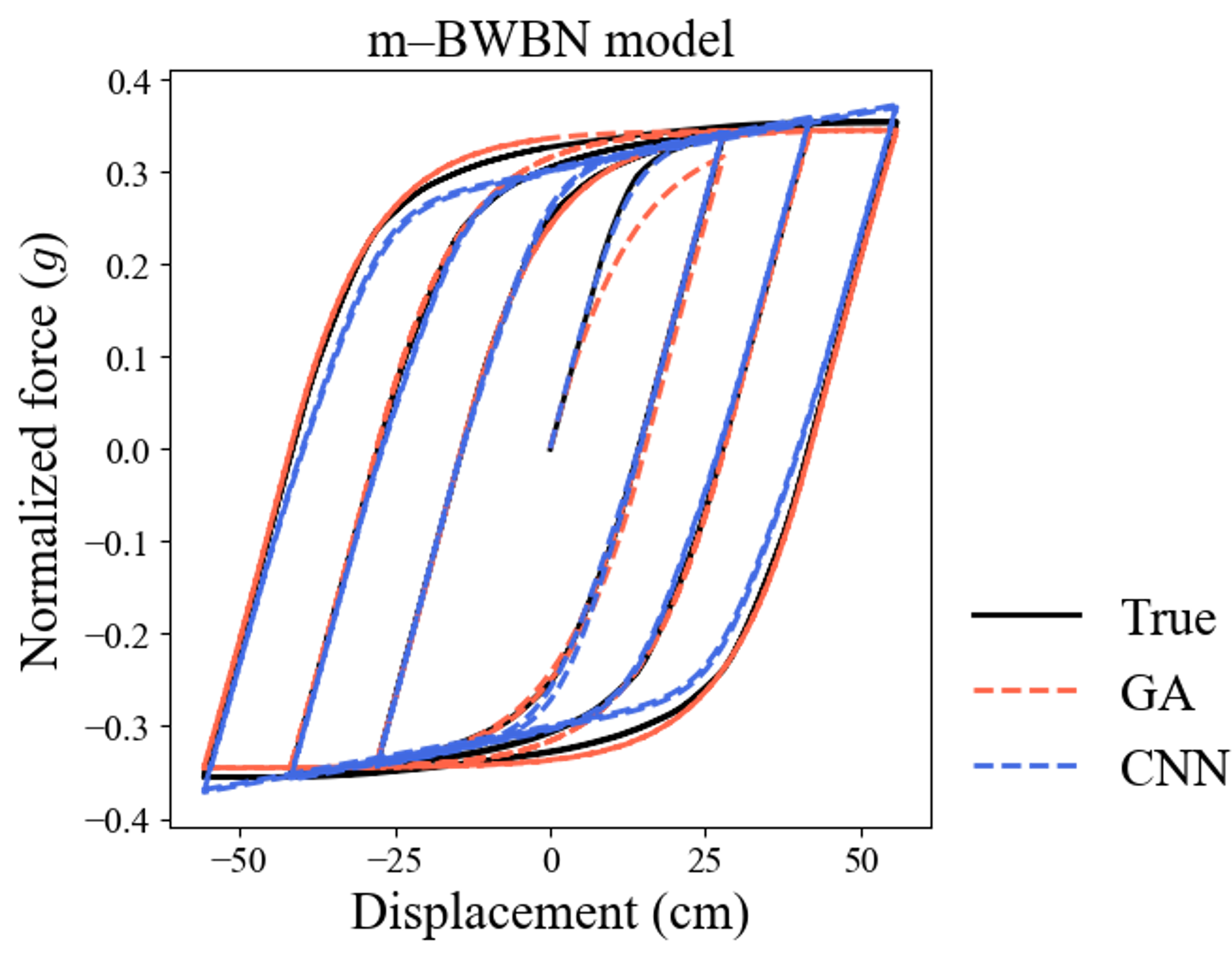}
            \caption{\textbf{}}
            \label{fig:Example1_hys_mBWBN}
        \end{subfigure}
        \vspace{-0.1in}
        \caption{\textbf{Comparison between the hysteresis curves for the (a) BW and (b) m-BWBN models, obtained using the OpenSees (True), the genetic algorithm (GA), and the CNN models (CNN). The corresponding optimal loading histories are used, as shown in Figure \ref{fig:optimal_loading_history}.}}
        \label{fig:Example1_hys}
    \end{figure}

    Figure \ref{fig:Example1_hys_BW} illustrates the parameter estimation results for the BW model using the two different methods. The close alignment for both methods illustrates the effectiveness of the proposed loading protocol. Notably, the hysteresis curve obtained using the genetic algorithm shows considerable deviation from the true curve at the beginning of the initial cycles, while the CNN-based curve closely follows the true curve. This exemplifies the limitations of genetic algorithm-based parameter estimation, where the fitness function is typically defined as the unweighted average of the error between true and predicted force values. On the other hand, the CNN-based estimation predicts the parameters directly from the spatial patterns, avoiding the issues associated with the fitness function.

    To further demonstrate the parameter estimation results, a nonlinear time history analysis is performed using the 1940 El Centro earthquake ground motion. The results are summarized in Table \ref{table:Example1_summary}. The table shows that the BW model estimated using the genetic algorithm significantly underestimates the maximum displacement, with a relative error exceeding 20\%, while the CNN-based BW model predicts the maximum displacement with a relative error of about 13\%. It is noted that the errors above 10\% observed in all cases are primarily attributable to differences in model fidelity between the high-fidelity OpenSees model and the equivalent single-degree-of-freedom system using BW class models. Still, for the genetic algorithm-based models, the substantial underestimation is observed, which is due to the underestimation of the force for the initial cycle, as discussed in Figure \ref{fig:Example1_hys_BW}. In addition, the reliable performance and the significant reduction in computational time of the CNN-based method emphasize the effectiveness of the proposed loading protocol.

    \begin{table}[H]
    \caption{\textbf{Comparison between the maximum displacement prediction results using the genetic algorithm-based (GA) and CNN-based (CNN) parameter estimation.}}
    \label{table:Example1_summary}
    \centering
        \begin{tabular}{lllll}
        \toprule[1pt]
        \textbf{Model}          & \textbf{Method}   & \textbf{Elapsed time} & {\textbf{\begin{tabular}[c]{@{}l@{}}Maximum\\displacement\end{tabular}}}    & {\textbf{\begin{tabular}[c]{@{}l@{}}Relative\\error\end{tabular}}}    \\
        \midrule[1pt]
        OpenSees                & -                 & -                     & 5.46 cm                          & -       \\
        \midrule
        \multirow{2}{*}{BW}     & GA                & 19.883 s              & 4.11 cm                          & 24.75\% \\
                                & CNN               & 0.283 s               & 4.73 cm                          & 13.37\% \\
        \midrule
        \multirow{2}{*}{m-BWBN} & GA                & 1162.742 s            & 4.29 cm                          & 21.33\% \\
                                & CNN               & 0.547 s               & 4.74 cm                          & 13.06\% \\
        \bottomrule[1pt]
        \end{tabular}
    \end{table}
    
    Note that we assumed the prior knowledge that the BW model would be sufficient to represent the hysteretic behavior of the target structure. To demonstrate a more general practice, another BW class model, the m-BWBN model, is further used to apply the modularized loading protocol. Based on the $u_y$ value determined as 14 cm in Figure \ref{fig:Example1_pushover}, the optimal loading history for the m-BWBN, shown in Figure \ref{fig:optimal_loading_history}, is constructed. The details of the genetic algorithm-based parameter estimation are the same as for the BW model. For the CNN-based parameter estimation, three independent CNN models are used to predict all BSC, DGD, and PCH parameters in the m-BWBN model. The alignment with the true hysteresis curve is shown in Figure \ref{fig:Example1_hys_mBWBN}, and the maximum displacement prediction results are summarized in Table \ref{table:Example1_summary}, both demonstrating the consistently reliable effectiveness of the proposed loading protocol, regardless of the specific BW class model. In fact, the DGD and PCH parameters estimated from both the genetic algorithm and CNN models are close to zero, making the estimated m-BWBN model essentially similar to the BW model. Hence, even if no prior knowledge is available, the proposed loading protocol can identify the appropriate parameter sets for the m-BWBN model, and the increase in computational time due to the use of a more complex model is negligible with CNN-based parameter estimation.

    \subsection{3-story RC frame}
    \noindent
    This section compares fragility curves for a structural system modeled using the BW class model constructed through the proposed protocol with those obtained from a high-fidelity OpenSees model. To this end, 3-story moment-resisting reinforced concrete (RC) frameis modeled using OpenSees. The details of the structure and the OpenSees model configuration are available in \citet{Kwon2006TheStructure}.

\begin{figure}[H]
        \centering
        \includegraphics[height=6cm]{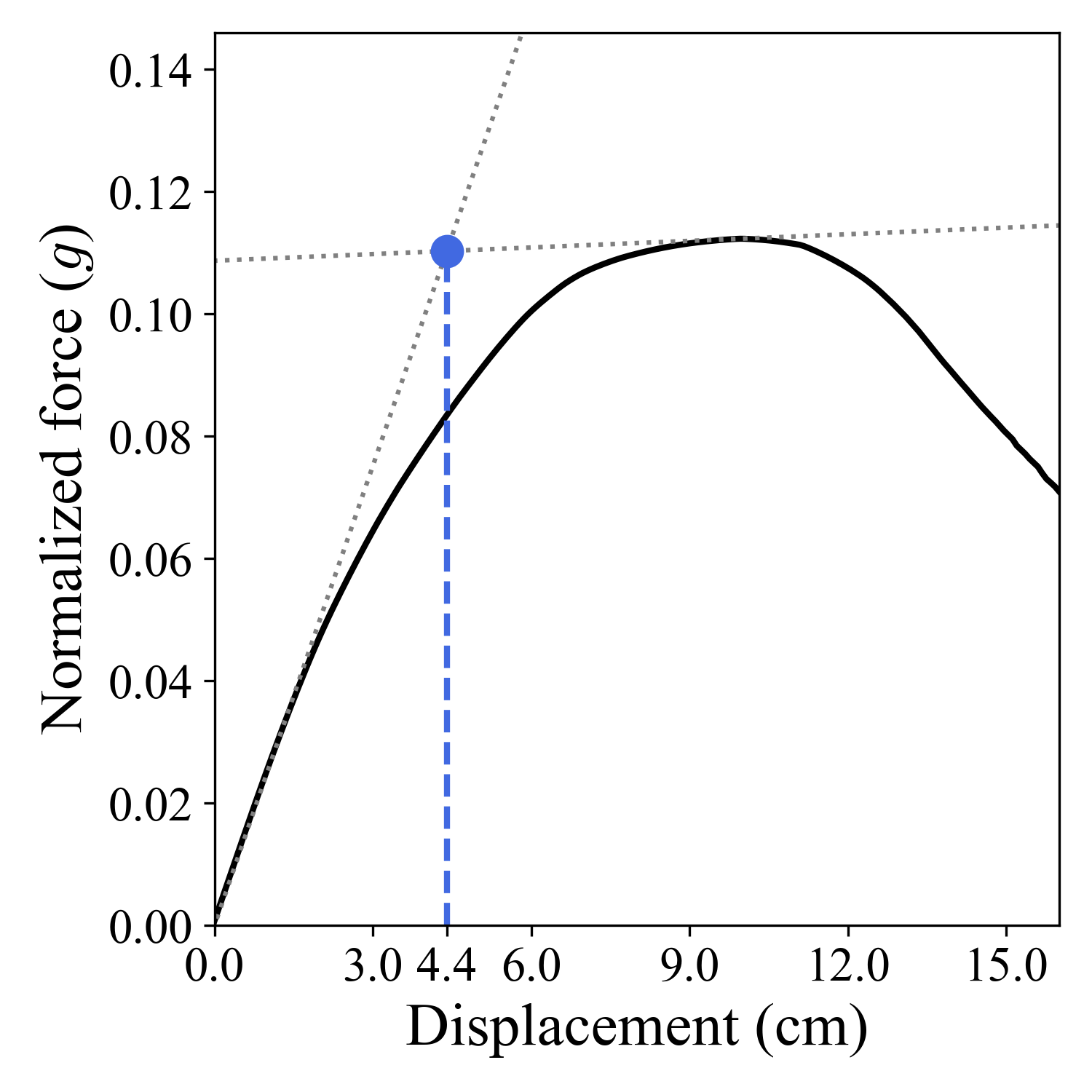}
        \vspace{-0.1in}
        \caption{\textbf{A pushover curve for the 3-story RC structure. The yield displacement $u_y$ is obtained as 4.4 cm.}}
        \label{fig:Example2_pushover}
    \end{figure}

    As illustrated in Figure \ref{fig:Example2_pushover}, the yield displacement is determined to be 4.4 cm. Next, since the RC structural system typically involves both structural degradation and the pinching effect, the optimal loading history for the m-BWBN model in Figure \ref{fig:optimal_loading_history} is employed to generate the hysteresis curve. The m-BWBN model parameters are then estimated from the generated hysteresis curve using a genetic algorithm and the three trained CNN models. 

\begin{figure}[H]
        \centering
        \includegraphics[height=6cm]{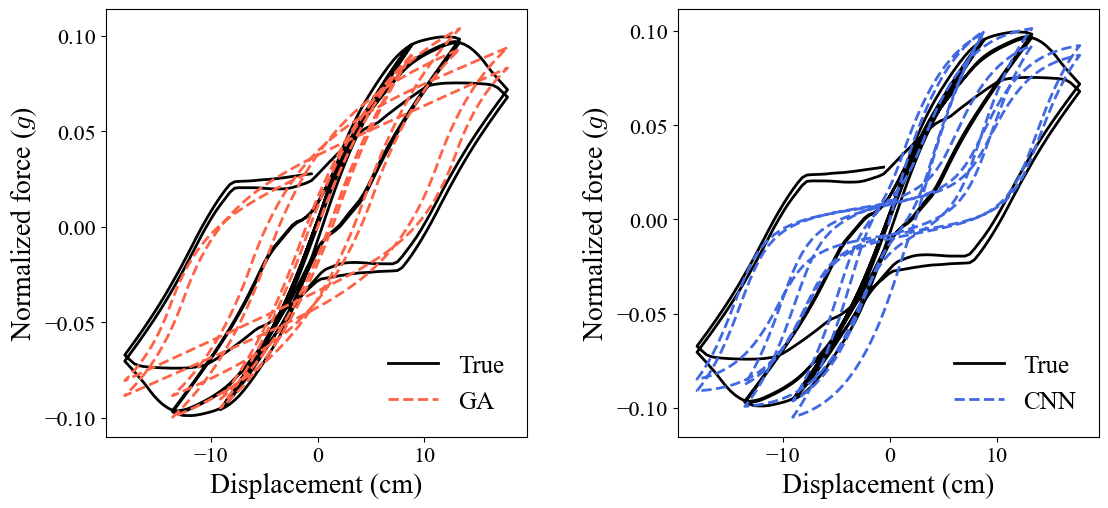}
        \vspace{-0.1in}
        \caption{\textbf{Comparison of the hysteresis curve obtained from the OpenSees (True) with the curves using the estimated parameters from the genetic algorithm (left) and the CNN models (right).}}
        \label{fig:Example2_hys}
    \end{figure}

    Figure \ref{fig:Example2_hys} compares the hysteresis curve from OpenSees with the curves obtained using the parameters estimated by the two different methods. The fitness of the predicted hysteresis curves to the OpenSees curve is limited compared to the case of the 3-story SAC structure in Figure \ref{fig:Example1_hys}, where the BW model adequately represents the hysteretic behavior. This discrepancy is partially attributed to the limitations of the m-BWBN model in accurately capturing the hysteretic behavior of this specific structure.

\begin{figure}[H]
        \centering
        \includegraphics[height=6cm]{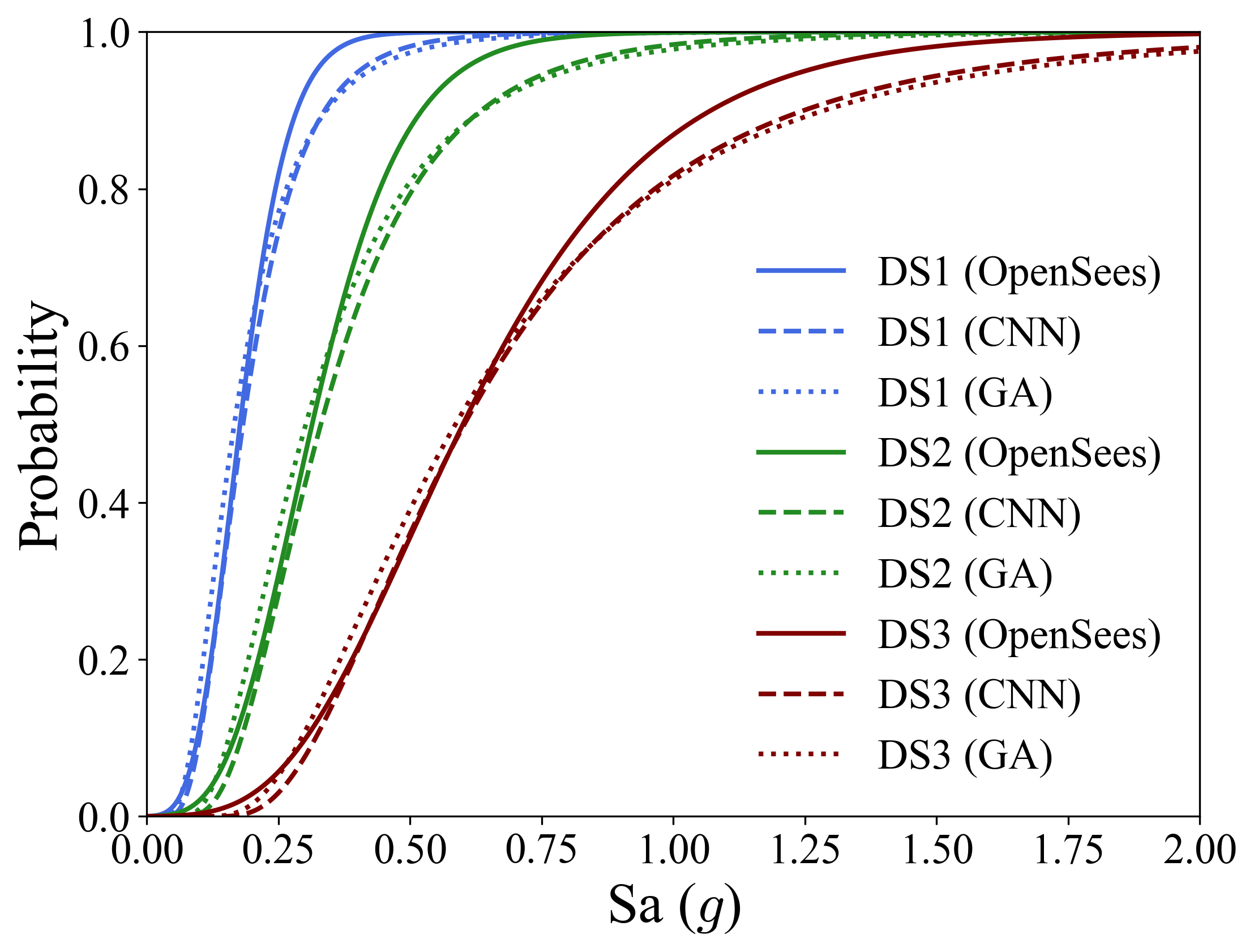}
        \vspace{-0.1in}
        \caption{\textbf{Fragility curves for three different damage states (DSs) obtained using the OpenSees and the m-BWBN model with the parameters estimated from the CNN models (CNN) and the genetic algorithm (GA).}}
        \label{fig:Example2_fragility}
    \end{figure}

    Based on the identified parameters, incremental dynamic analyses are performed to construct fragility curves using a total of 135 ground motions. The ground motion time histories are collected from the NGA-West database \cite{Chiou2008NGADatabase, Power2008AnProject}. Three damage states of the structure are defined by roof displacements of 50 mm (DS1), 90 mm (DS2), and 170 mm (DS3) following \citet{Kim2020ProbabilisticMethod}. The fragility curve obtained from OpenSees is compared with those of a single-degree-of-freedom system equipped with the m-BWBN model using two different parameter sets: one estimated from the genetic algorithm and the other from the CNN models. These comparisons are depicted in Figure \ref{fig:Example2_fragility} and summarized in Table \ref{table:Example2_summary}.
    
    The fragility curves derived using the m-BWBN model, whether through the genetic algorithm or CNN models, closely match the OpenSees-based fragility curve across all three damage states, demonstrating the overall effectiveness of the proposed loading protocol for the m-BWBN model. The closeness of the fragility curves is further confirmed by the Kullback-Leibler (KL) divergence in Table \ref{table:Example2_summary}. The KL divergence is a metric that measures the distance between two probability distributions, with values closer to zero indicating greater alignment \cite{Kullback1951OnSufficiency}. The gaps observed at relatively large $S_a$ values are partly due to the limitations of representing a high-fidelity structural model (OpenSees) with a low-fidelity model (equivalent single-degree-of-freedom system using the m-BWBN model). However, these gaps do not significantly reduce the effectiveness of the overall protocol, especially in analyses involving a large number of structures where microscopic details become less significant. Lastly, the elapsed time in Table \ref{table:Example2_summary} describes the total time for both parameter estimation and incremental dynamic analysis, highlighting the remarkable efficiency of the CNN-based method; it takes approximately 500 times and 2,000 times less computational time compared to OpenSees- and genetic algorithm-based fragility curve construction, respectively.

    \begin{table}[H]
    \caption{\textbf{Comparison between the genetic algorithm-based (GA) and the CNN models-based (CNN) fragility curves for three damage states (DSs).}}
    \label{table:Example2_summary}
    \centering
        \begin{tabular}{lllll}
        \toprule[1.0pt]
        \multirow{2}{*}[-0.6ex]{\textbf{Method}}    & \multicolumn{3}{c}{\textbf{KL divergence}}    & \multirow{2}{*}[-0.6ex]{\textbf{\begin{tabular}[c]{@{}l@{}}Total\\elapsed time\end{tabular}}} \\ \cmidrule{2-4}
                                                    & DS1       & DS2       & DS3                   &                                                                                               \\
        \midrule[1.0pt]
        OpenSees                                    & -         & -         & -                     & 273,515.96 s                                                                                   \\
        \midrule
        GA                                          & 0.1281    & 0.1342    & 0.1015                & 2,378.85 s + 374.71 s                                                                          \\
        \midrule
        CNN                                         & 0.0811    & 0.1140    & 0.1458                & 3.02 s + 516.08 s                                                                             \\
        \bottomrule[1.0pt]
        \end{tabular}
    \end{table}

    \section{Conclusions}\label{section:conclusions}
    \noindent
    A deep learning-based modularized loading protocol was proposed for reliable and efficient parameter estimation for Bouc-Wen (BW) class models. The protocol consists of two key components: optimal loading history and the convolutional neural network (CNN)-based rapid parameter estimation. By leveraging the ability of CNNs to capture spatial correlations between data points, we developed CNN architectures that effectively capture the dependencies between parameters and hysteresis behavior. After training these architectures on 18 different loading histories, minimal loading histories (referred to as loading history modules) were identified for each of the three hysteresis categories (basic hysteresis, structural degradation, and the pinching effect), which provide estimation performance comparable to the reference loading history. These modules can be combined to construct optimal loading histories for different BW class models. Finally, the protocol, which includes yield displacement determination, optimal loading history construction, and rapid parameter estimation using CNN models, was proposed with example optimal loading histories. Each of the three CNN models for parameter estimation and the three loading history modules for an optimal loading history can be used either collectively or independently, making the proposed protocol modular and adaptable to the specific hysteresis categories being modeled.
    
    The numerical investigations demonstrated the efficiency and effectiveness of the protocol, but some limitations remain. The lower accuracy in estimating pinching parameters compared with other hysteretic characteristics suggests the need for further refinement. Ongoing research focuses on developing a physics-based neural network that incorporates the equation of motion with the BW class model as a loss function, which could address these challenges. Additionally, extending the protocol to diverse hysteresis models beyond the BW class models, covering a broader range of hysteretic behaviors and structural systems, could further enhance its applicability and robustness in real-world applications.

    \section*{Acknowledgement}
    \noindent
    The second author is supported by Institute of Construction and Environmental Engineering at Seoul National University. The third author is supported by the National Research Foundation of Korea (NRF) grant funded by the Korea government (MSIT) (RS-2023-00242859).

    \bibliography{references}

    \appendix
    
    \section{Training and test errors for CNN models}\label{appendix:mse_errors}
    \noindent
    Table \ref{table:appendix_cnn_architectures} summarizes the training and test errors for the CNN models trained on the reference loading history (denoted as Ref. in the table), providing a detailed demonstration of the effectiveness of the developed CNN architectures. In addition, Table \ref{table:appendix_cnn_optimal} details the errors for the CNN models trained on the optimal loading histories shown in Figure \ref{fig:optimal_loading_history}, where $\text{Opt}_{\text{BW}}$, $\text{Opt}_{\text{BWdeg}}$, and $\text{Opt}_{\text{m-BWBN}}$ denote the optimal loading histories for the BW, BW  with degradation, and m-BWBN models. These three CNN models are used as rapid parameter estimators in the proposed loading protocol.
    
    \begin{table}[H]
    \caption{\textbf{Training and test errors for the CNN models based on the CNN architectures developed in Section \ref{section:cnn_architectures}.}}
    \label{table:appendix_cnn_architectures}
    \centering
        \begin{tabular}{>{\raggedright\arraybackslash}m{1.8cm} >{\raggedright\arraybackslash}m{3cm} >{\raggedright\arraybackslash}m{2cm} >{\centering\arraybackslash}m{1.8cm} >{\centering\arraybackslash}m{2.2cm} >{\centering\arraybackslash}m{2.2cm}}
        \toprule[1.0pt]
        \multicolumn{3}{c}{\textbf{CNN Model information}} & \multirow{2}{*}[-0.8em]{\textbf{Parameter}} & \multicolumn{2}{c}{\textbf{Mean squared error}} \\ 
        \cmidrule(lr){1-3} \cmidrule(lr){5-6}
        \textbf{Target category} & \textbf{Trained loading history} & \textbf{Related Section} & & \textbf{Training} & \textbf{Test} \\ 
        \midrule[1.0pt]
        \multirow{5}{*}{\textbf{BSC}} & \multirow{5}{*}{Ref.} & \multirow{5}{*}{Section \ref{section:cnn_architectures_BSC_DGD}} & \centering $T$ & $2.935 \times 10^{-4}$ & $5.119 \times 10^{-4}$ \\ 
         &  & & \centering $F_y$ & $3.169 \times 10^{-4}$ & $9.933 \times 10^{-4}$  \\ 
         &  & & \centering $\alpha$ & $1.239 \times 10^{-5}$ & $1.560 \times 10^{-5}$  \\ 
         &  & & \centering $\beta$ & $6.028 \times 10^{-4}$ & $1.421 \times 10^{-3}$  \\ 
         &  & & \centering $n$ & $1.030 \times 10^{-2}$ & $1.895 \times 10^{-2}$  \\ 
        \midrule
        \multirow{2}{*}{\textbf{DGD}} & \multirow{2}{*}{Ref.} & \multirow{2}{*}{Section \ref{section:cnn_architectures_BSC_DGD}} & \centering $\delta_\nu$ & $1.074 \times 10^{-4}$ & $6.100 \times 10^{-4}$ \\
         &  &  & \centering $\delta_\eta$ & $1.116 \times 10^{-4}$ & $2.613 \times 10^{-4}$ \\ 
        \midrule
        \multirow{7}{*}{\textbf{PCH}} & \multirow{7}{*}{Ref.} & \multirow{7}{*}{Section \ref{section:cnn_architectures_PCH}} & \centering $\zeta_0$ & $1.246 \times 10^{-2}$ & $1.067 \times 10^{-1}$ \\ 
         &  &  & \centering $p$ & $3.747 \times 10^{-2}$ & $2.271 \times 10^{-1}$\\ 
         &  &  & \centering $q$ & $1.478 \times 10^{-3}$ & $1.825 \times 10^{-2}$\\ 
         &  &  & \centering $\psi$ & $2.429 \times 10^{-2}$ & $9.648 \times 10^{-2}$\\ 
         &  &  & \centering $\delta_\psi$ & $2.726 \times 10^{-4}$ & $1.537 \times 10^{-3}$\\
         &  &  & \centering $\lambda$ & $4.255 \times 10^{-2}$ & $1.732 \times 10^{-1}$\\ 
        \bottomrule[1.0pt]
        \end{tabular}
    \end{table}

    \begin{table}[H]
    \caption{\textbf{Training and test errors for CNN models trained on the optimal loading histories for three BW class models.}}
    \label{table:appendix_cnn_optimal}
    \centering
        \begin{tabular}{>{\raggedright\arraybackslash}m{1.8cm} >{\raggedright\arraybackslash}m{3cm} >{\raggedright\arraybackslash}m{2cm} >{\centering\arraybackslash}m{1.8cm} >{\centering\arraybackslash}m{2.2cm} >{\centering\arraybackslash}m{2.2cm}}
        \toprule[1.0pt]
        \multicolumn{3}{c}{\textbf{CNN Model information}} & \multirow{2}{*}[-0.8em]{\textbf{Parameter}} & \multicolumn{2}{c}{\textbf{Mean squared error}} \\ 
        \cmidrule(lr){1-3} \cmidrule(lr){5-6}
        \textbf{Target category} & \textbf{Trained loading history} & \textbf{Related Section} & & \textbf{Training} & \textbf{Test} \\ 
        \midrule[1.0pt]
        \multirow{5}{*}{\textbf{BSC}} & \multirow{5}{*}{$\text{Opt}_{\text{BW}}$} & \multirow{5}{2cm}{Sections \ref{section:loading_history_bsc} and \ref{section:proposed_loading_protocol}} & \centering $T$ & $2.307 \times 10^{-4}$ & $1.124 \times 10^{-3}$ \\ 
         &  & & \centering $F_y$ & $6.027 \times 10^{-5}$ & $2.263 \times 10^{-4}$  \\ 
         &  & & \centering $\alpha$ & $2.821 \times 10^{-5}$ & $3.276 \times 10^{-5}$  \\ 
         &  & & \centering $\beta$ & $4.869 \times 10^{-4}$ & $1.303 \times 10^{-3}$  \\ 
         &  & & \centering $n$ & $7.178 \times 10^{-3}$ & $3.609 \times 10^{-2}$  \\ 
        \midrule
        \multirow{2}{*}{\textbf{DGD}} & \multirow{2}{*}{$\text{Opt}_{\text{BWdeg}}$} & \multirow{2}{2cm}{Sections \ref{section:loading_history_dgd} and \ref{section:proposed_loading_protocol}} & \centering $\delta_\nu$ & $5.511 \times 10^{-4}$ & $1.923 \times 10^{-4}$ \\
         &  &  & \centering $\delta_\eta$ & $1.260 \times 10^{-4}$ & $2.870 \times 10^{-4}$ \\ 
        \midrule
        \multirow{6}{*}{\textbf{PCH}} & \multirow{6}{*}{$\text{Opt}_{\text{m-BWBN}}$} & \multirow{6}{2cm}{Sections \ref{section:loading_history_pch} and \ref{section:proposed_loading_protocol}} & \centering $\zeta_0$ & $1.868 \times 10^{-2}$ & $1.049\times 10^{-1}$\\ 
         &  &  & \centering $p$ & $1.402 \times 10^{-2}$ & $2.495 \times 10^{-1}$\\ 
         &  &  & \centering $q$ & $1.387 \times 10^{-3}$ & $9.524 \times 10^{-3}$\\ 
         &  &  & \centering $\psi$ & $2.200 \times 10^{-2}$ & $7.521 \times 10^{-2}$\\ 
         &  &  & \centering $\delta_\psi$ & $3.909 \times 10^{-4}$ & $1.448 \times 10^{-3}$\\
         &  &  & \centering $\lambda$ & $3.166 \times 10^{-2}$ & $1.155 \times 10^{-1}$\\ 
        \bottomrule[1.0pt]
        \end{tabular}
    \end{table}

    \section{Histograms for generated m-BWBN model parameters}\label{appendix:histograms}
\begin{figure}[H]
        \centering
        \includegraphics[width=1.0\linewidth]{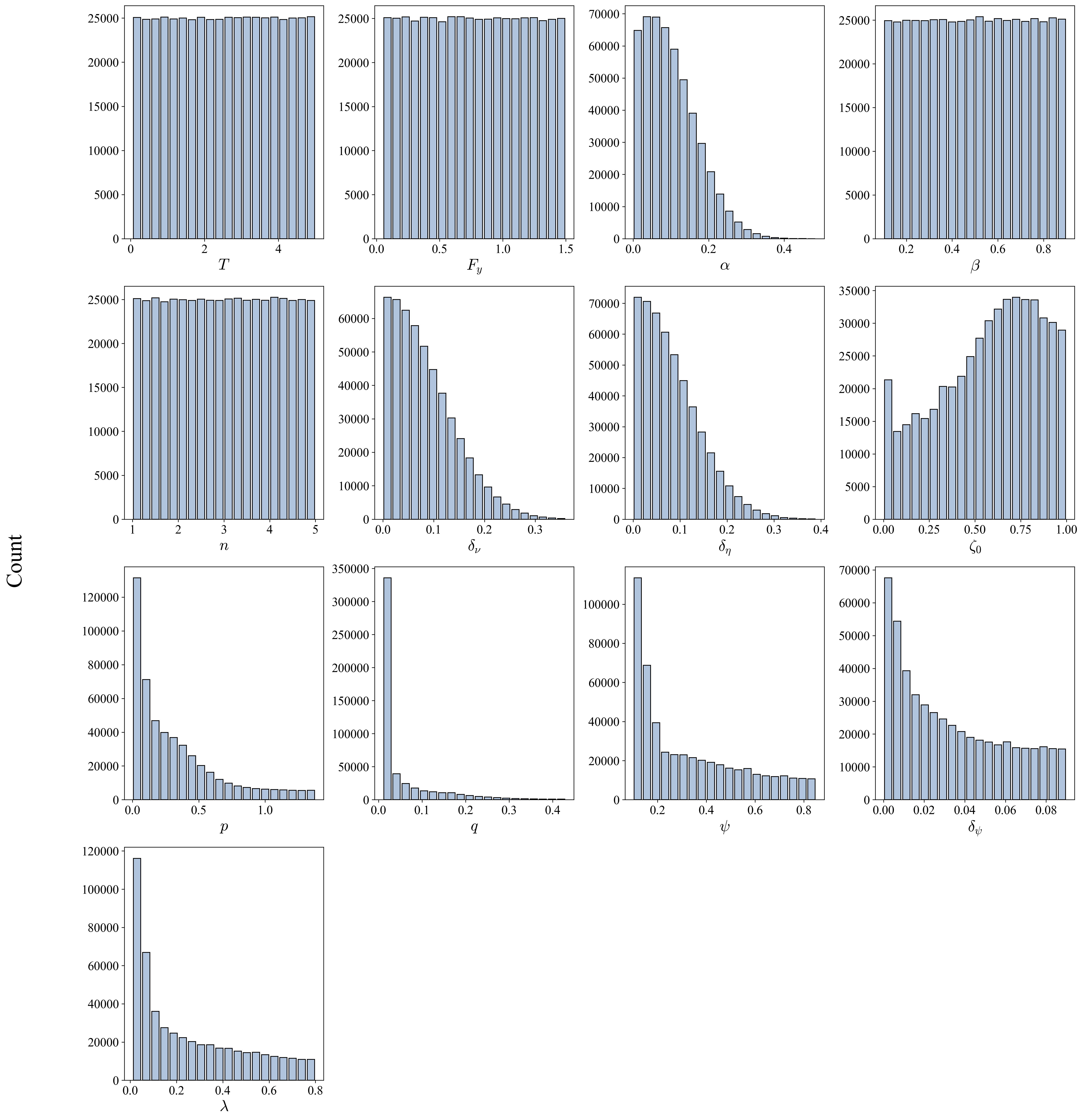}
        \vspace{-0.2in}
        \caption{\textbf{Histograms of the generated samples for each parameter.}}        
        \label{fig:param_hist}
    \end{figure}

    \noindent
    Figure \ref{fig:param_hist} presents the histograms of the 500,000 m-BWBN model parameters generated as discussed in Section \ref{section:data_generation}.

    \section{Scatter plots comparing optimal and reference loading histories}\label{appendix:scatter_plots}
\begin{figure}[H]
    \centering
    \begin{subfigure}{0.95\textwidth}
        \centering
        \includegraphics[width=0.8\textwidth]{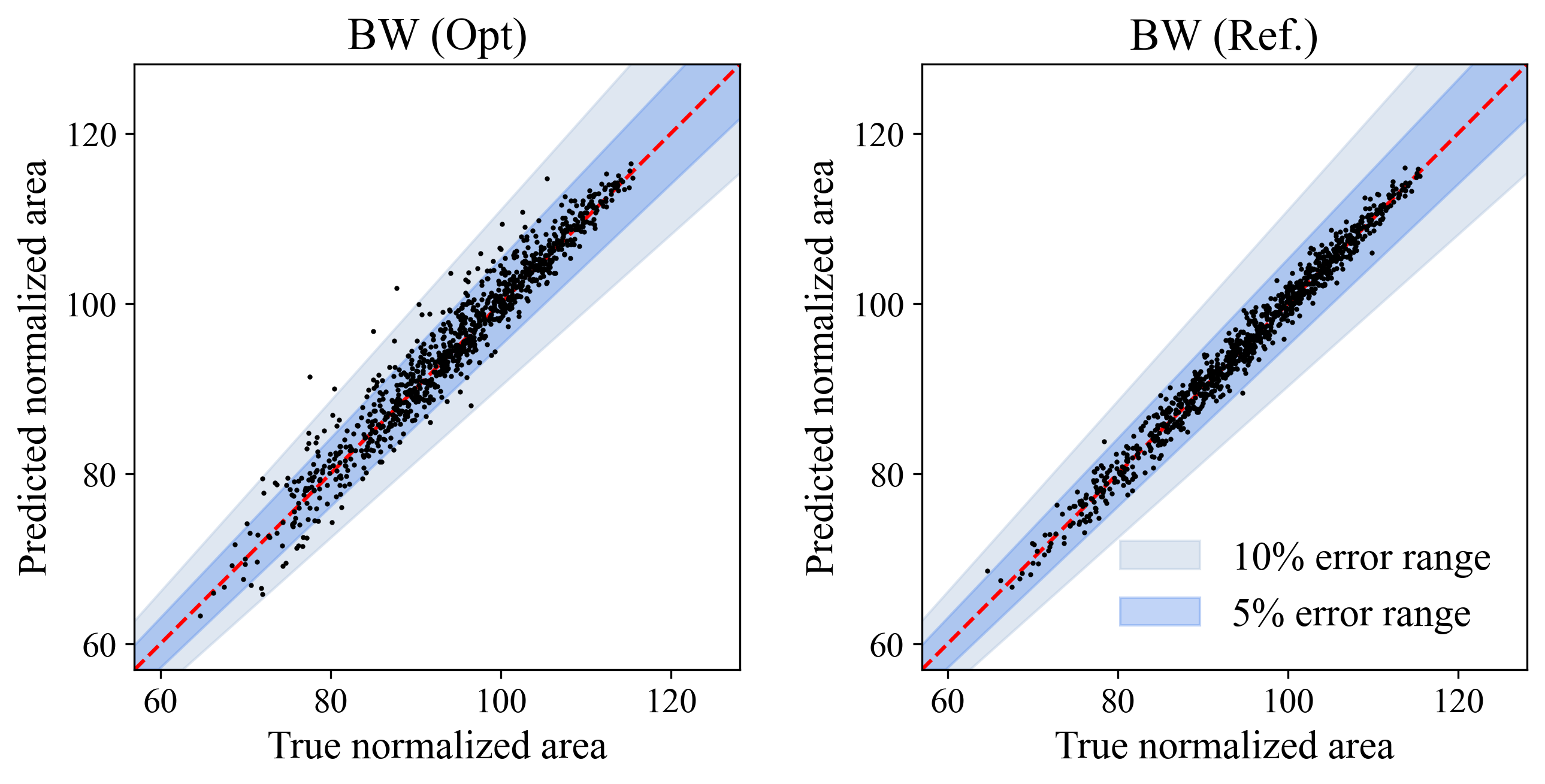}
        \caption{\textbf{}}
        \label{fig:LH_demo_energy_BW}
    \end{subfigure}
    \begin{subfigure}{0.95\textwidth}
        \centering
        \includegraphics[width=0.8\textwidth]{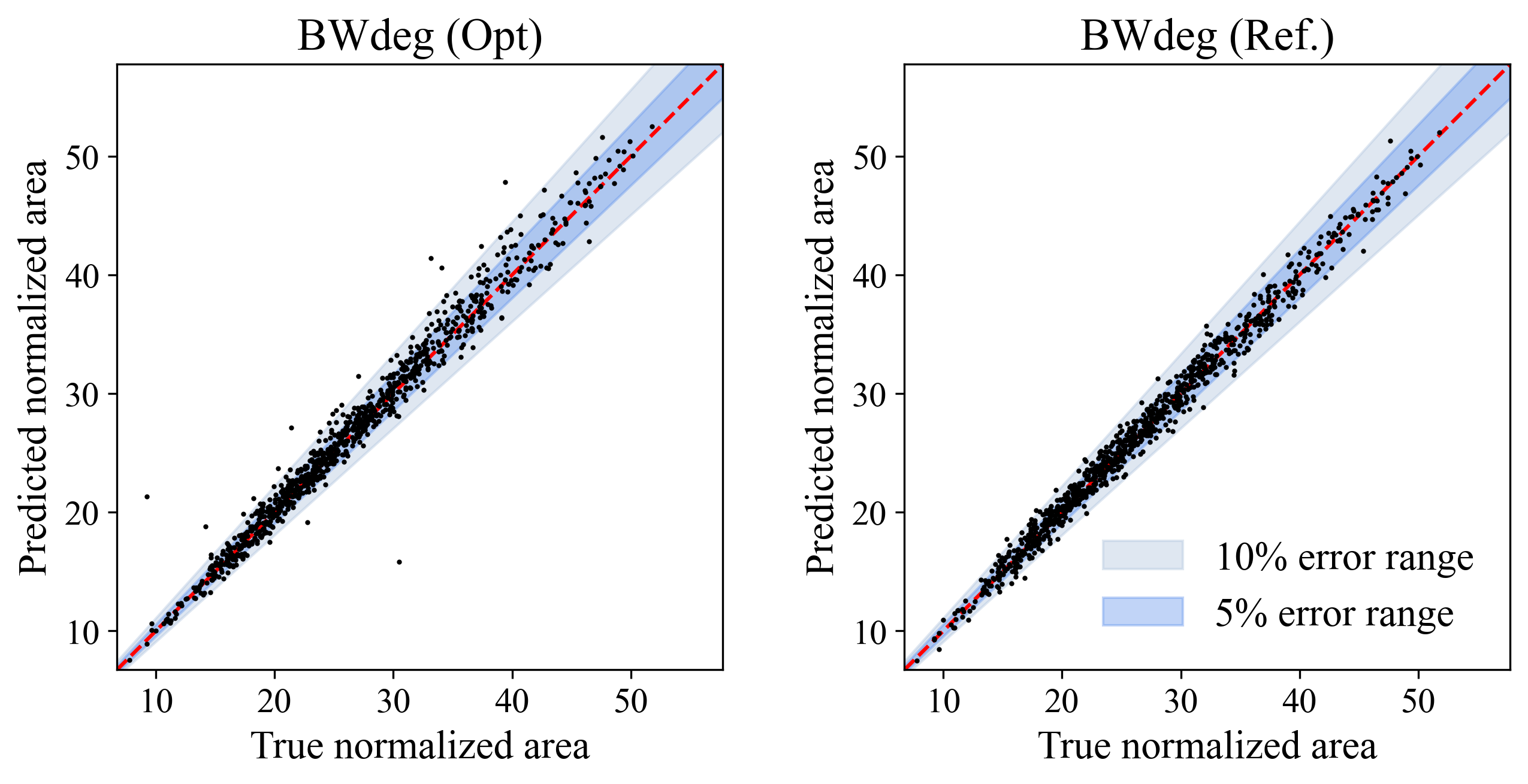}
        \caption{\textbf{}}
        \label{fig:LH_demo_energy_BWdeg}
    \end{subfigure}
    \begin{subfigure}{0.95\textwidth}
        \centering
        \includegraphics[width=0.8\textwidth]{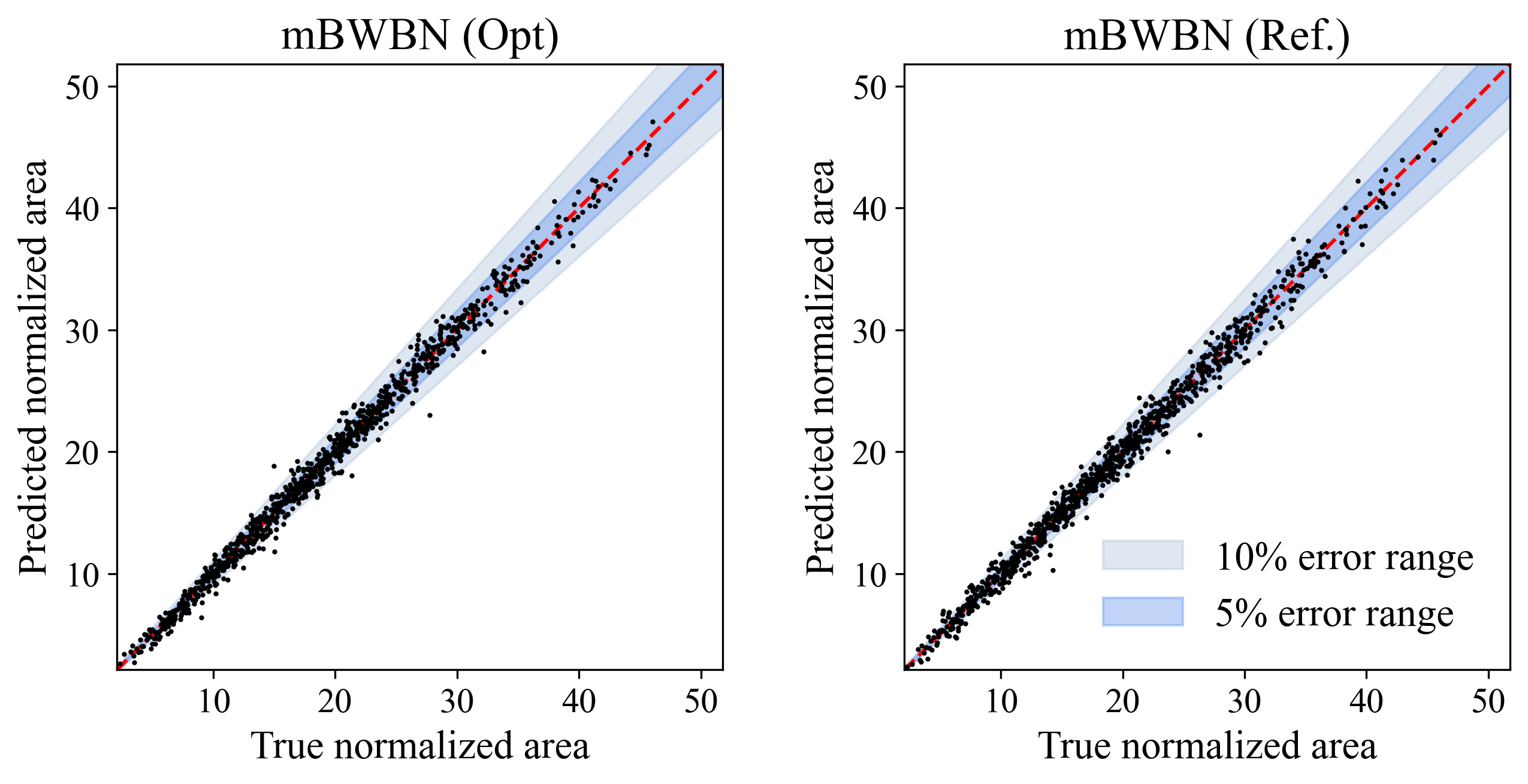}
        \caption{\textbf{}}
        \label{fig:LH_demo_energy_mBWBN}
    \end{subfigure}
    \caption{\textbf{Comparison between the true area of a hysteresis curve with the area predicted using the optimal loading history (Opt) and the reference loading history (Ref.) for (a) the BW model, (b) the BW model with degradation, and (c) the m-BWBN model.}}         
    \label{fig:LH_demo_energy}
    \end{figure}

    \noindent
    Figure \ref{fig:LH_demo_energy} illustrates the scatter plots comparing the true area under a hysteresis curve with the area predicted using the optimal and reference loading histories, where the area is normalized by the yield strength $F_y$ and the yield displacement $u_y$. Each black dot represents each 1,000 data point, and the light blue and plain blue regions indicate the error margins of 10\% and 5\%, respectively. The red dashed line denotes the identity line.
    
\end{document}